\documentclass[review]{elsarticle}
\graphicspath{ {./figures/} }
\usepackage{hyperref}
\usepackage{float}
\usepackage{verbatim} %comments
\usepackage{apalike}
\usepackage{amsmath,amssymb,amsfonts}
\usepackage{algorithmic}
\usepackage{graphicx}
\usepackage[table,xcdraw]{xcolor}
\usepackage{multirow}
\usepackage{array}
\usepackage{booktabs}
\usepackage{caption}
\usepackage{textcomp}
\usepackage{enumitem}
\usepackage{tabularx}

\usepackage[export]{adjustbox}
\usepackage{amsfonts}
\usepackage{algorithmic}
\usepackage[ruled,vlined,linesnumbered,boxruled]{algorithm2e}
\usepackage{setspace}

\newcolumntype{C}[1]{>{\centering\let\newline\\\arraybackslash\hspace{0pt}}m{#1}}

\restylefloat{figure}
\restylefloat{table}

\journal{Engineering Applications of Artificial Intelligence}

\biboptions{authoryear}

\begin{document}
\begin{frontmatter}

\begin{titlepage}
\begin{center}
\vspace*{1cm}

\textbf{ \large Domain-knowledge Inspired Pseudo Supervision (DIPS) for Unsupervised Image-to-Image Translation Models to Support Cross-Domain Classification}

\vspace{0.5cm}

% Author names and affiliations
Firas Al-Hindawi$^{a}$ (falhinda@asu.edu), Md Mahfuzur Rahman Siddiquee$^a$ (mrahmans@asu.edu), Teresa Wu$^a$ (teresa.wu@asu.edu), Han Hu$^b$ (hanhu@uark.edu),  Ying Sun$^c$ (sunyg@ucmail.uc.edu) \\

% \hspace{10pt}

\section*{Highlights}
% \item Research highlight 1
% \item Research highlight 2
\begin{itemize}

\item New metric for evaluating UI2I translation models using pseudo supervised metrics that is designed specifically to support cross-domain classification.

% \item Gaussian Mixture Model is utilized to generate pseudo labels to enable the use of standard supervised metrics. 

\item The metric not only outperforms unsupervised metrics such as the FID, but is also highly correlated with the true supervised metrics, robust, and explainable.

\item The applicability of the metric in this research field is demonstrated through a critical real-world application (the boiling crisis problem). 

\end{itemize}

\begin{flushleft}
\small  
% $^a$ Full address of first author, including the country name \\
% $^b$ Full address of second author, including the country name \\
% $^c$ Full address of last author, including the country name
$^a$ Arizona State University, 699 Mill Avenue, Tempe, AZ 85281, US\\
$^b$ University of Arkansas, 1 University of Arkansas, Fayetteville, AR 72701, US\\
$^c$ University of Cincinnati, 2600 Clifton Ave, Cincinnati, OH 45221, US\\

\begin{comment}
Clearly indicate who will handle correspondence at all stages of refereeing and publication, also post-publication. Ensure that phone numbers (with country and area code) are provided in addition to the e-mail address and the complete postal address. Contact details must be kept up to date by the corresponding author.
\end{comment}

\vspace{1cm}

\textbf{Corresponding Author:} \\
Firas Al-Hindawi \\
Arizona State University, 699 Mill Avenue, Tempe, AZ 85281, US \\
Tel: +1 (602) 837-9820 \\
Email: falhinda@asu.edu

\end{flushleft}        
\end{center}
\end{titlepage}

\title{Domain-knowledge Inspired Pseudo Supervision (DIPS) for Unsupervised Image-to-Image Translation Models to Support Cross-Domain Classification}

\author[label1]{Firas Al-Hindawi \corref{cor1}}
\ead{falhinda@asu.edu}

\author[label1]{Md Mahfuzur {Rahman Siddiquee}}
    \ead{mrahmans@asu.edu}

\author[label1]{Teresa Wu}
    \ead{teresa.wu@asu.edu}

\author[label2]{Han Hu}
    \ead{hanhu@uark.edu}
    
\author[label3]{Ying Sun}
    \ead{sunyg@ucmail.uc.edu}

\cortext[cor1]{Corresponding author.}
\address[label1]{Arizona State University, 699 Mill Avenue, Tempe, AZ 85281, US}
\address[label2]{University of Arkansas, 1 University of Arkansas, Fayetteville, AR 72701, US}
\address[label3]{University of Cincinnati, 2600 Clifton Ave, Cincinnati, OH 45221, US}

\begin{abstract}
The ability to classify images is dependent on having access to large labeled datasets and testing on data from the same domain of which the model was trained on. Classification becomes more challenging when dealing with new data from a different domain, where gathering and especially labeling a larger image dataset for retraining a classification model requires a labor-intensive human effort. Cross-domain classification frameworks were developed to handle this data domain shift problem by utilizing unsupervised image-to-image translation models to translate an input image from the unlabeled domain to the labeled domain. The problem with these unsupervised models lies in their unsupervised nature. For lack of annotations, it is not possible to use the traditional supervised metrics to evaluate these translation models to pick the best-saved checkpoint model. This paper introduces a new method called  Domain-knowledge Inspired Pseudo Supervision (DIPS) which utilizes Gaussian Mixture Models and domain knowledge to generate pseudo annotations to enable the use of traditional supervised metrics. This method was designed specifically to support cross-domain classification applications contrary to other typically used metrics such as the Fréchet Inception Distance (FID) which were designed to evaluate the model in terms of the quality of the generated image from a human-eye perspective. DIPS outperforms state-of-the-art GAN evaluation metrics when selecting the optimal saved checkpoint. Furthermore, DIPS showcases its robustness and interpretability by demonstrating a strong correlation with truly supervised metrics, highlighting its superiority over existing state-of-the-art alternatives. The boiling crisis problem has been approached as a case study. The code and data to replicate the results can be found on the official \href{https://github.com/Hindawi91/DIPS}{GitHub repository}\footnote{\url{https://github.com/Hindawi91/DIPS}}.

\end{abstract}

\begin{keyword}
Unsupervised Metrics \sep  Cross-Domain Classification \sep Critical Heat Flux \sep Domain Adaptation \sep Generative Adversarial Networks \sep Image-to-Image Translation \sep Pool Boiling \sep Unsupervised Machine Learning.
\end{keyword}

\end{frontmatter}

\section{Introduction}
\label{introduction}

Machine learning prediction and classification algorithms have been a rapidly growing field in recent years, particularly with the significant advancements in deep learning and computer vision technologies. These advancements enabled prediction algorithms to become highly efficient and accurate, making them applicable to a wide range of applications in various domains such as tensile strength prediction of polymers \citep{alhindawi2018predicting,altarazi2019machine}, Critical heat flux detection \citep{rassoulinejad-mousavi2021a,al2023framework}, soil fertility classification \citep{padmapriya2023deep}, and skin lesion classification \citep{omeroglu2023novel}. For classification models to work optimally, there are several contingencies to consider, one of which is having access to large, balanced, and accurately labeled datasets.\\

The limitations of these classification models become evident when dealing with new data from a different domain. In such situations, gathering a substantial dataset with labels and creating a new classifier from the beginning may require significant time and resources which may not always be practical, this problem is known in the field as the data domain shift problem. This prompted the development of a branch of machine learning called unsupervised domain adaptation (UDA), which deals with the data domain shift problem in an unsupervised manner. The specific classification problem described earlier falls under the UDA umbrella and is generally referred to in the field of machine learning as unsupervised cross-domain classification, which is the problem of training a classifier on a dataset from one domain and using it to predict a dataset from a different domain without the labeling information. UDA could be categorized into discrepancy-based, reconstruction-based, or adversarial-based UDA depending on the domain adaptation approach used \citep{wang2018deep}. Researchers used a variety of these UDA approaches to tackle the cross-domain classification problem, such as transforming the original image into analogs in multiple related domains, thereby learning features that are robust to variations across domains \citep{ghifary2015domain}, and aligning the distributions of the domains using extracted features \citep{tzeng2017adversarial}, or multiple features representations \citep{zhu2019multi}  to capture the information from different aspects. Such methods can handle domains with big shifts between them, but in the case of domains with small shifts, the lost spatial information during transformation might affect the results negatively. \\

Recently, with the rise of Generative Adversarial Networks (GANs) and Unsupervised Image to Image (UI2I) translation models, researchers started investigating their utility in solving the cross-domain classification problem. Image-to-image translation models, in general, are used to convert an input image from one domain (e.g., horse images) to another domain (e.g., zebra images) using a generative model. Once trained, UI2I translation models can be used to transform an input image from the source domain to the target domain by inputting the image into the model and generating a synthetic output image in the target domain. This translation capability inspired a number of approaches that leverage GANs and UI2I models to address the cross-domain classification problem.\\ 

For example, \cite{deng2018image} translated the source dataset to the target domain and then trained a new model on the features of the translated images. \cite{xiang2020unsupervised} synthesized a dataset and fine-tuned the model using the synthesized dataset. \cite{li2021cross} utilized GANs and attention to leverage the information available from the source domain to the target domain. \cite{al2023framework} developed a framework using UI2I translation models that expanded the classification capability to include the target domain. \cite{goel2023unsupervised} Proposed a framework that utilizes a guided transfer learning approach to select layers for fine-tuning, enhancing feature transferability, and minimizing domain discrepancies using JS-Divergence.\\

One of the main challenges in such unsupervised frameworks is knowing which UI2I model to select/save during training (when to stop the training). If the model was not trained long enough, it will underfit the training data and produce poor results. In contrast, if the model was overtrained, it will overfit the training data and generate poor results as well. Because of the unsupervised nature of UI2I models, the standard supervised metrics that are used to validate supervised models during training (e.g., accuracy, AUC) are not available to be used without access to the labeling information. To overcome this issue, such frameworks depend on one of the most used metrics in evaluating unsupervised image-to-image translation models, the Frechet Inception Distance (FID) \cite{heusel2017a}. FID is a popular metric for evaluating the quality of generated images in the context of Generative Adversarial Networks (GANs) and other image synthesis models. However, like any metric, FID has its drawbacks and may not always be the best choice for evaluating the performance of a particular model \citep{borji2022pros}. Some of the general disadvantages of FID are that the Gaussian assumption in FID calculation might not hold in practice, the FID has high bias, and the sample size to calculate FID has to be large enough (usually above 50k), otherwise it would lead to an over-estimation of the actual FID \citep{borji2022pros}. Moreover, it is computationally expensive \citep{https://doi.org/10.48550/arxiv.2009.14075}. For the specific application of unsupervised cross-domain classification such as the framework mentioned in \citep{al2023framework}, the FID was helpful in selecting a relatively good model, but it had problems. Most of the time it wasn’t able to pick the best possible model; moreover, the FID model ranking was not correlated with the true model ranking based on the true supervised classification metrics, making the model selection choice seem random and unexplainable. It was stressed in \citep{al2023framework} that there is a need for a framework to properly assess the validation datasets in order to improve the UI2I translation model selection criteria in unsupervised cross-domain classification models. The reason why traditional GAN evaluation metrics such as the FID are not suitable to support cross-domain classification is that they were not designed for that purpose, but rather were designed to evaluate the model in terms of the quality of the generated image from a human-eye perspective, and its ability to generate diverse results (not falling into mode collapse). Moreover, they do not take advantage of prior domain knowledge available in classification tasks such as the number of classes expected.\\ 

This work proposes a framework for evaluating UI2I translation models designed to support cross-domain classification applications using pseudo-supervised metrics. In this proposed approach, the Inception model is utilized to extract the features from the unlabeled target Dataset, but unlike other methods such as the Inception Score (IS) or the FID, this approach utilizes an unsupervised clustering technique known as the Gaussian Mixture Models to create pseudo labels which enable the use of standard supervised metrics. This methodology is shown not only to outperform unsupervised metrics such as the FID, but also is highly correlated with the true supervised metrics and mimics the monotonically decreasing behavior of their model ranking. This demonstrates the robustness and explainability of the metric, unlike the FID which is poorly correlated with the true supervised metrics and has inconsistent ranking order that is neither robust nor explainable. To showcase the efficiency of the methodology, the boiling crisis detection problem was used as an example. The boiling crisis, also known as critical heat flux (CHF), represents a formidable challenge in the field of thermal engineering and is of paramount importance to various industrial processes, particularly in nuclear reactors, electronics cooling, and power generation systems \citep{rassoulinejad-mousavi2021a}. This phenomenon occurs when the rate of heat transfer from a heated surface to a boiling liquid abruptly deteriorates, leading to a sudden increase in surface temperatures and potentially catastrophic consequences \citep{rassoulinejad-mousavi2021a}. Resolving the boiling crisis is critical as it can prevent the formation of vapor film insulating the surface, thereby averting equipment damage, reactor meltdowns, and ensuring the safety and efficiency of numerous technological applications \citep{rassoulinejad-mousavi2021a}, making it a vital and urgent research frontier. Recent efforts were dedicated to alleviating this problem using machine learning techniques \cite{rassoulinejad-mousavi2021a,rokoni2022a} and most recently \cite{al2023framework} developed a cross-domain classification framework using UI2I translation models to tackle this problem. Their work serves as a suitable case study for the method proposed in this manuscript, especially since using an unreliable metric such as FID was one of the limitations. The same datasets used by the authors in \cite{al2023framework} were also used in this work. Two experiments were conducted using the two publicly available datasets ($DS1$ and $DS2$) by alternating the target and source datasets for each experiment ($DS1  \rightarrow{} DS2$ and $DS2  \rightarrow{} DS1$).

To summarize the contribution of this manuscript:

\begin{itemize}
  \item This work introduces a new framework for evaluating UI2I translation models using pseudo-supervised metrics. The framework was designed specifically to support cross-domain classification.
  \item The introduced framework utilizes an unsupervised clustering technique (GMM) to cluster the extracted features into N clusters, where N is the number of classes known as a prior from domain knowledge, and use these clusters as pseudo labels to enable the use of standard supervised metrics. 
  \item The framework not only outperforms unsupervised metrics such as the FID, but is also highly correlated with the true supervised metrics, robust, and explainable.
\end{itemize}

The rest of the manuscript is organized as follows, Section 2 discusses relevant GAN-based I2I translation studies and related evaluation frameworks. Section 3 describes the proposed framework and the analysis procedures used in the study. Section 4 details the conducted experiments and Section 5 showcases and discusses the results. Finally, the manuscript is concluded in section 6, followed by a listing of the sources cited afterward.

\section{Related Work} \label{RelatedWork}

This section is composed of two parts. The first part will briefly discuss the evolution of GANs in the literature and expand on their utilization in both supervised and unsupervised image-to-image translation. The second part discusses the most common GAN evaluation measures and compares their advantages and disadvantages.

\subsection{GANs and Image-to-Image Translation}

First introduced by \cite{goodfellow2014a}, GANs consist of two networks, a generator and a discriminator that are trained together for an overall objective of generating realistic synthetic data that resembles a given dataset. The generator takes a random noise vector as input and transforms it into synthetic data that aims to resemble the real dataset. The discriminator evaluates both the real and the fake generated data samples and assigns scores indicating their authenticity. During training, the generator seeks to maximize these scores, while the discriminator strives to correctly distinguish between real and fake data. This adversarial process continues iteratively until the generator generates data that is indistinguishable from real data, resulting in GANs capable of generating high-quality synthetic data samples \citep{goodfellow2014a}.

The development of GANs has been the subject of numerous research papers and has led to the introduction of various variations and extensions of the original GAN architecture, such as Conditional GANs \citep{mirza2014conditional}, InfoGANs \citep{chen2016infogan}, Progressive GANs \citep{karras2017progressive}, and Wasserstein GANs \citep{arjovsky2017wasserstein}. The first work to utilize GANs to solve the I2I translation problem was \citep{isola2017image}. In their pix2pix model, they used conditional adversarial networks to learn the mapping from an input image to an output image, where the networks learn a loss function to train this mapping. This method uses a "U-Net" based architecture for the generator and a "PatchGAN" classifier for the discriminator. Multiple efforts were spent to improve and build upon the pix2pix model and overcome its weaknesses. The major pitfall of the pix2pix model was that it was supervised. The training process required  paired images in the training set for the model to learn the mapping $G: X \rightarrow Y$. Thus, to solve this problem, \cite{zhu2017toward} introduced CycleGAN. The authors coupled the adversarial loss with an inverse mapping $F: Y \rightarrow X$ and introduced a cycle consistency loss. The objective of this loss is to enforce $F(G(X)) \approx X$ and $G(F(Y)) \approx Y$. A similar approach was performed by \citep{yi2017dualgan} and \citep{kim2017learning} concurrently with cycleGAN. Building on these works,\cite{Choi_2018_CVPR} introduced their StarGAN framework that simultaneously trains multiple datasets with different domains using a single generator and discriminator pair. However, StarGAN tends to change the images unnecessarily during image-to-image translation even when no translation is required \citep{Siddiquee_2019_ICCV}. To address this issue, \cite{Siddiquee_2019_ICCV} proposed the Fixed-Point GAN (FP-GAN) framework. This framework focused on identifying a minimal subset of pixels for domain translation and introduced fixed-point translation by supervising same-domain translation through a conditional identity loss and regularizing cross-domain translation through revised adversarial, domain classification, and cycle consistency losses.

\subsection{GANs Evaluation Metrics}

With the rapid rise of GANs and their use in various applications, the need for evaluation metrics to assess these models became increasingly critical. Traditional image evaluation measures such as the peak-signal-to-noise ratio (PSNR) and the structural similarity index measure (SSIM) were mainly designed to support tasks related to image compression and restoration where the image-to-image similarity was of utmost importance. Thus, they focused on measuring the similarity between images and were not suitable for image synthesis tasks. Depending on the application, there are two main types of GAN evaluation measures currently used, qualitative and quantitative measures. In this section, the focus will be on quantitative measures since the proposed work falls under that category. Currently, the most common quantitative GAN evaluation measures are the IS \cite{salimans2016improved}, the FID \cite{heusel2017a}, the Maximum Mean Discrepancy (MMD) \cite{gretton2008kernel} and the Kernel Inception Distance (KID) \cite{binkowski2018demystifying}. This section discusses these metrics and Table \ref{metrics_summary} summarizes their strengths and weaknesses.

\begin{table}[H]
\centering
\scriptsize
\caption{Strengths and weaknesses of common GAN evaluation metrics.}
\label{metrics_summary}

\begin{tabular}{|>{\centering\arraybackslash}m{1.15cm}|>{\centering\arraybackslash}m{4.7cm}|>{\centering\arraybackslash}m{4.7cm}|}
\hline
\textbf{Metric} & \textbf{Strengths} & \textbf{Weaknesses} \\
\hline

MMD & 
\begin{itemize}[partopsep=0pt,parsep=0pt,itemsep=0pt,leftmargin=*]
  \item Versatile metric applicable to various data distributions.
  \item Not Dependent on Inception model.
  \item Captures mode collapse and diversity.
\end{itemize} & 
\begin{itemize}[partopsep=0pt,parsep=0pt,itemsep=0pt,leftmargin=*]
  \item Choice of kernel function and hyperparameters influence performance.
  \item Computationally expensive.
  \item Not designed for cross-domain classification.
\end{itemize} \\
\hline
KID & 
\begin{itemize}[partopsep=0pt,parsep=0pt,itemsep=0pt,leftmargin=*]
  \item Captures quality and diversity.
  \item Less sensitive to Inception model.
  \item More interpretable than FID.
\end{itemize} & 
\begin{itemize}[partopsep=0pt,parsep=0pt,itemsep=0pt,leftmargin=*]
  \item Depends on Inception model.
  \item Computationally expensive.
  \item Not designed for cross-domain classification.
\end{itemize} \\
\hline
IS & 
\begin{itemize}[partopsep=0pt,parsep=0pt,itemsep=0pt,leftmargin=*]
  \item Encourages visually appealing and diverse images.
  \item Captures quality and diversity.
  
\end{itemize} & 
\begin{itemize}[partopsep=0pt,parsep=0pt,itemsep=0pt,leftmargin=*]
  \item Biased towards ImageNet.
  \item Sensitive to model parameters.
  \item Requires a large sample size.
  \item Insensitive to intra-class diversity
  \item Not interpretable.
  \item Does not compare source images with target images.
  \item Not designed for cross-domain classification.
\end{itemize} \\
\hline
FID & 
\begin{itemize}[partopsep=0pt,parsep=0pt,itemsep=0pt,leftmargin=*]
  \item Correlates with human judgment.
  \item Captures quality and diversity.
  \item Relatively stable and consistent.
  \item Can detect intra-class mode collapse.
\end{itemize} & 
\begin{itemize}[partopsep=0pt,parsep=0pt,itemsep=0pt,leftmargin=*]
  \item Depends on Inception model.
  \item Not interpretabile.
  \item Not designed for cross-domain classification.
  \item Requires a large sample size.
  \item Biased towards ImageNet.
\end{itemize} \\
\hline

\end{tabular}

\end{table}

 \subsubsection{Maximum Mean Discrepancy (MMD)}

The Maximum Mean Discrepancy (MMD) is a statistical measure used to quantify the discrepancy between two probability distributions \citep{wynne2022kernel}. It provides a way to assess the dissimilarity between samples drawn from different distributions and is commonly used in machine learning and generative modeling to compare the distributions of real and generated data samples \citep{wilson2016deep,li2015generative}. The MMD measure is based on the idea of comparing the means of feature representations of samples from each distribution \citep{wynne2022kernel}. By calculating the MMD, it can be determined whether two sets of data come from the same distribution or differ significantly \citep{wynne2022kernel}. The general formula for MMD can be found in Table \ref{Metrics}. In this equation, $MMD_u^2(X, Y)$ represents the squared unbiased MMD statistic between $X$ and $Y$. The variables $m$ and $n$ represent the numbers of samples in distributions $X$ and $Y$, respectively. The variables $x_i$ and $y_i$ represent individual samples from distributions $X$ and $Y$, respectively. $k$ denotes the kernel function applied to samples from two distributions. The equation consists of three main terms. The first term calculates the average kernel similarity between all pairs of samples within distribution $X$. The second term calculates the average kernel similarity between all pairs of samples within distribution $Y$. The third term calculates the average kernel similarity between samples from distribution $X$ and distribution $Y$. By comparing these three terms, the MMD quantifies the discrepancy or difference between the distributions $X$ and $Y$.

Main advantages of MMD is the ease of implementation and the rich kernel based theory behind it, making it a versatile metric that can be applied to various types of data distributions. It is less dependent on the specific architecture of the feature extractor and can also capture both the mode collapse and diversity of generated samples. It is however sensitive to the choice of kernel function and can be computationally expensive for large datasets.

% can be given as follows:

% \begin{equation}
% \begin{aligned}
% MMD_u^2(X, Y) &= \frac{1}{m(m-1)} \sum_{i \neq j}^m k\left(x_i, x_j\right) \\
% &\quad + \frac{1}{n(n-1)} \sum_{i \neq j}^n k\left(y_i, y_j\right) \\
% &\quad - \frac{2}{m n} \sum_{i=1}^m \sum_{j=1}^n k\left(x_i, y_j\right)
% \end{aligned}
% \label{MMD_equation}
% \end{equation}

\subsubsection{Kernel Inception Distance (KID)}

The KID metric is a method used for evaluating the quality and diversity of generated images in the field of generative adversarial networks (GANs). The KID utilizes the MMD metric introduced earlier to measure the discrepancy between the features extracted from real and generated images using the InceptionV3 neural network. The general mathematical expression of the KID is listed in Table \ref{Metrics}. where $f_{real}$ and $f_{fake}$ represent the extracted features from real and fake images using the inception model.

% Equation \ref{KID_equation} shows the general mathematical expression of the KID: 

% \begin{equation}
% \text {KID} = \text{MMD}\left(f_{\text {real }}, f_{\text {fake }}\right)^2
% \label{KID_equation}
% \end{equation}

KID has multiple advantages. It enables a more objective evaluation of GAN performance and it considers both the intra-class and inter-class variations of features, capturing the high-level semantics of images. Moreover, KID is model-agnostic and can be applied to evaluate GANs trained with different architectures and datasets. However, the KID also suffers from a number of disadvantages. It assumes that the Inception network's feature representations are sufficient to capture the quality and diversity of images even though it may not capture all aspects of image quality. It also does not capture lower-level image characteristics, such as pixel-level details and spatial coherence; because it used the extracted features in it's calculations. Furthermore, the choice of the kernel function in KID can heavily influence the metric.

\subsubsection{Inception Scores (IS)}

The IS Eq. listed in Table \ref{Metrics}. is a measure of the quality and diversity of generated images, based on the KL divergence between the predicted and true class distributions of a pre-trained Inception network \citep{borji2022pros,treder2022quality}. where $\mathbb{E}_{x \sim p_{\mathrm{gen}}}$ is the expectation over the images sampled from the generator, $\mathbb{K} \mathbb{L}$ refers to the Kullback-Leibler divergence, $p(y)$ is the marginal distribution of class labels, and $(p(y \mid x)$ represents the conditional distribution of class labels $y$ given an input image $x$.

% \begin{equation}
% \mathrm{IS}=\exp \mathbb{E}_{x \sim p_{\mathrm{gen}}}(\mathbb{K} \mathbb{L}(p(y \mid x) \| p(y)))
% \label{InceptionScore}
% \end{equation}

Although used commonly, The IS has some downsides. The IS does not capture intra-class diversity and lacks correlation with the human judgment of image quality and may produce inconsistent results when compared to human evaluations\citep{barratt2018note}. Moreover, it is insensitive to the prior distribution over labels (hence is biased towards ImageNet dataset and Inception model), and is very sensitive to model parameters and implementations \citep{borji2022pros}). The IS also requires a large sample size to be reliable.

\subsubsection{Fréchet Inception Distance (FID)}

Similar to IS, the FID depends on the Inception model to generate its value, the difference, however, is that the FID calculates the Wasserstein-2 (a.k.a Fréchet) distance between multivariate Gaussians fitted to the embedding space of the Inception-v3 network of generated and real images \citep{borji2022pros,treder2022quality}. The general mathematical expression of the FID is listed in Table \ref{Metrics}. In this equation, $\mu_g$ and $\mu_r$ represent the means of the feature maps of the generated images and real images, respectively, and $C_g$ and $C_r$ represent the covariance matrices of the feature maps of the generated images and real images, respectively. $Tr$ represents the trace operator, which sums the diagonal elements of a matrix. Unlike IS, the FID is more consistent with human inspection, is sensitive to minimal changes in the real distribution, and can detect intra-class mode collapse \citep{borji2022pros}. That being said, the FID has shortcomings as well. For example, the Gaussian assumption in FID calculation is not always valid, the FID has high bias, and requires a large sample size ($\geq 50k$ images) to be efficient \citep{chong2020effectively}.

% The FID is Denoted as:

% \begin{equation}
% \text{FID} = ||\mu_{g} - \mu_{r}||^{2} + Tr(C_{g} + C_{r} - 2(C_{g}C_{r})^{1/2})
% \label{FID}
% \end{equation}

% Despite these and the efforts of many others, these metrics either focus solely on the diversity of the results or the image quality from a human-eye perspective. There has yet to emerge a metric or framework that is designed to support the cross-domain classification applications of UI2I translation models.

\cite{siddiquee2023brainomaly} showed that the I2I model selected by FID has a weak correlation with the target classification task; therefore, the I2I model selected by FID performs poorly on the classification task. As a result, they proposed a pseudo-AUC metric for their anomaly detection task. Although this work was inspired by their work, their proposed pseudo-AUC metric cannot be applied directly to our task as they had images partially annotated in their problem setting. 

Despite these efforts and those of others, these metrics either focus solely on the diversity of the results, the image quality from a human-eye perspective or require a portion of the target domain to be partially annotated (as the case with the pseudo-AUC). A metric or framework designed to support fully unsupervised cross-domain classification frameworks has yet to emerge.

\section{Methodology} \label{Methodology}

\begin{figure}[H]
    \centering
    \includegraphics[width = \columnwidth, frame]{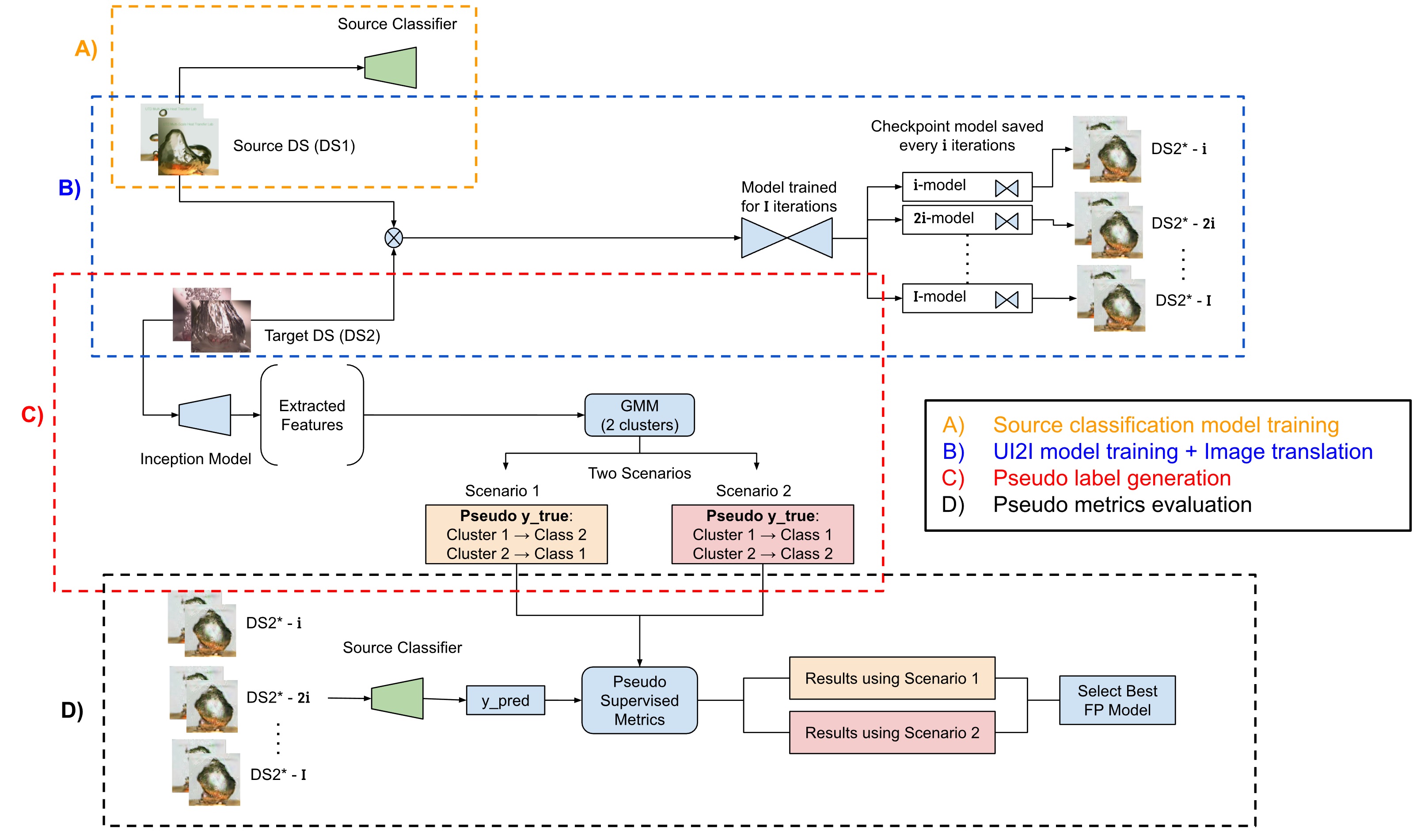}
    \caption{ Pseudo Metric Entire Framework: part A) in orange shows the classification model training, part B) shows the UI2I model training and the image translation, part C) shows the pseudo label generation and part D) shows the final pseudo metrics evaluation.}
    \label{fig:framework}
\end{figure}

This section describes the methodology behind the proposed metric. The entire framework is summarized in both Figure \ref{fig:framework} and Algorithm \ref{alg:Pseudo code}. As shown, the methodology is broken down into smaller parts and described in detail in each sub-section as follows. Step 1 in Algorithm \ref{alg:Pseudo code} (depicted in Part A) in Figure \ref{fig:framework}) represents the source classification model training, which is further explained in \ref{CNN_Training}. Steps 2 and 3 in Algorithm \ref{alg:Pseudo code} (depicted in Part B) in Figure \ref{fig:framework}) represent the UI2I model training and the cross-domain image translation process. Detailed explaination in sections \ref{UI2I_training} and \ref{image_translation} respectively. Finally, Steps 4 and 5 in Algorithm \ref{alg:Pseudo code} (depicted in Parts C and D) in Figure \ref{fig:framework}) explain how the pseudo labels were created and incorporated to generate the proposed pseudo-supervised metrics. Details are documented in section \ref{Pseudo_eval_sec}.

The demonstrated figures in this section show only one of the experiments, which is when $DS1$ is used as a source dataset and $DS2$ as a target dataset ($DS1 \xrightarrow{} DS2$). The same logic applies to the other experiment.

\begin{algorithm}[H]
\scriptsize 
\setstretch{0.6}
\caption{Pseudo Supervised Metrics Algorithm}

\label{alg:Pseudo code}
  \DontPrintSemicolon
    \SetKwFunction{one}{train\_source\_classifier}
    \SetKwFunction{two}{train\_UI2I\_model}
    \SetKwFunction{three}{translate\_target}
    \SetKwFunction{four}{generate\_pseudo\_labels}
    \SetKwFunction{five}{pseudo\_supervised\_metrics}
  
  \SetKwProg{Fn}{Function}{:}{}

 \textbf{ Step 1:}
  \Fn{\one{$source\_DS$}}{
        train model\;
        save best model\;
        source-DS classifier = best saved model\;
        \KwRet{source-DS classifier}\;
  }
  
  \textbf{Step 2:}
  \Fn{\two{$source\_DS$,$target\_DS$}}{
        \textbf{initialize} models\_list = [ ]\;
        use source\_DS and target\_DS to train UI2I translation model\;
        run training for $I$ iterations\;
        save a checkpoint model every $i$ iterations\;
        append saved checkpoint model to models\_list\;
        \KwRet{models\_list [$i$-model, $2i$-model, ... , $I$-model]}\;
  }
  \textbf{Step 3:}
    \Fn{\three{$target\_DS$,$models\_list$}}{
        \textbf{initialize} translated\_sets\_list = [ ]\;
        \textbf{For} each model in models\_list:\;
        \Indp translate target\_DS to source domain using model\;
        append translated images set to translated\_sets\_list\;
        \Indm \KwRet{translated\_sets\_list [$source\_DS^*-i$, $source\_DS^*-2i$, ... , $source\_DS^*-I$]}\;
  }
  \textbf{Step 4:}
    \Fn{\four{$target\_DS$,$GMM$,$Inception\_model$}}{
        extract features from the target\_DS\;
        use GMM to cluster the data into 2 clusters\;
        assign labels to clusters covering both possible scenarios:\;
        \Indp \textbf{Scenario 1:} \{cluster 1 = label 1, cluster 2 = label 2 \}\;
        \textbf{Scenario 2:} \{cluster 1 = label 2, cluster 2 = label 1 \}\;
        \Indm \KwRet{ Pseudo\_$y_{true}$(scenario 1), Pseudo\_$y_{true}$ (scenario 2)}\;
  }
  
  \textbf{Step 5:}
  \Fn{\five{$translated\_sets\_list$,Pseudo $y_{true}$,Source-DS classifier}}{
        \textbf{initialize} models\_results\_list = [ ]\;
        \textbf{For} each translated\_set in translated\_sets\_list:\;
        \Indp $y_{pred} =$ Source-DS\_classifier(translated\_set)\;
        generate Pseudo metrics using $y_{pred}$ and pseudo $y_{true}$ from scenario 1\;
        generate Pseudo metrics using $y_{pred}$ and pseudo $y_{true}$ from scenario 2\;
        append best Pseudo metrics results to models\_results\_list\;
        \Indm best\_UI2I\_model = best(models\_results\_list)\;
        \KwRet{best\_UI2I\_model}\;
  }

\end{algorithm}

\subsection{Source Classification Model Training}
\label{CNN_Training}
% Figure \ref{fig:CNN-training} summarizes our source classifier training approach. 
The source dataset is split into three subsets, training, validation, and testing. A classification model is then trained on the training set for a pre-set number of iterations and the model is validated after every epoch using the validation dataset. The model that scores the best on the validation dataset is saved. Afterward, the best-saved model is tested on the testing set for final evaluation. For the purpose of these experiments, a convolutional neural network (CNN) was used as a classifier, but the methodology is agnostic to the type of classification model used.

% \begin{figure}[H]
%     \centering
%     \includegraphics[width = \columnwidth, frame]{CNN training.jpg}
%     \caption{Source classifier training process}
%     \label{fig:CNN-training}
% \end{figure}

\subsection{UI2I model training process}
\label{UI2I_training}

\begin{figure}[H]
    \centering
    \includegraphics[width = \columnwidth, frame]{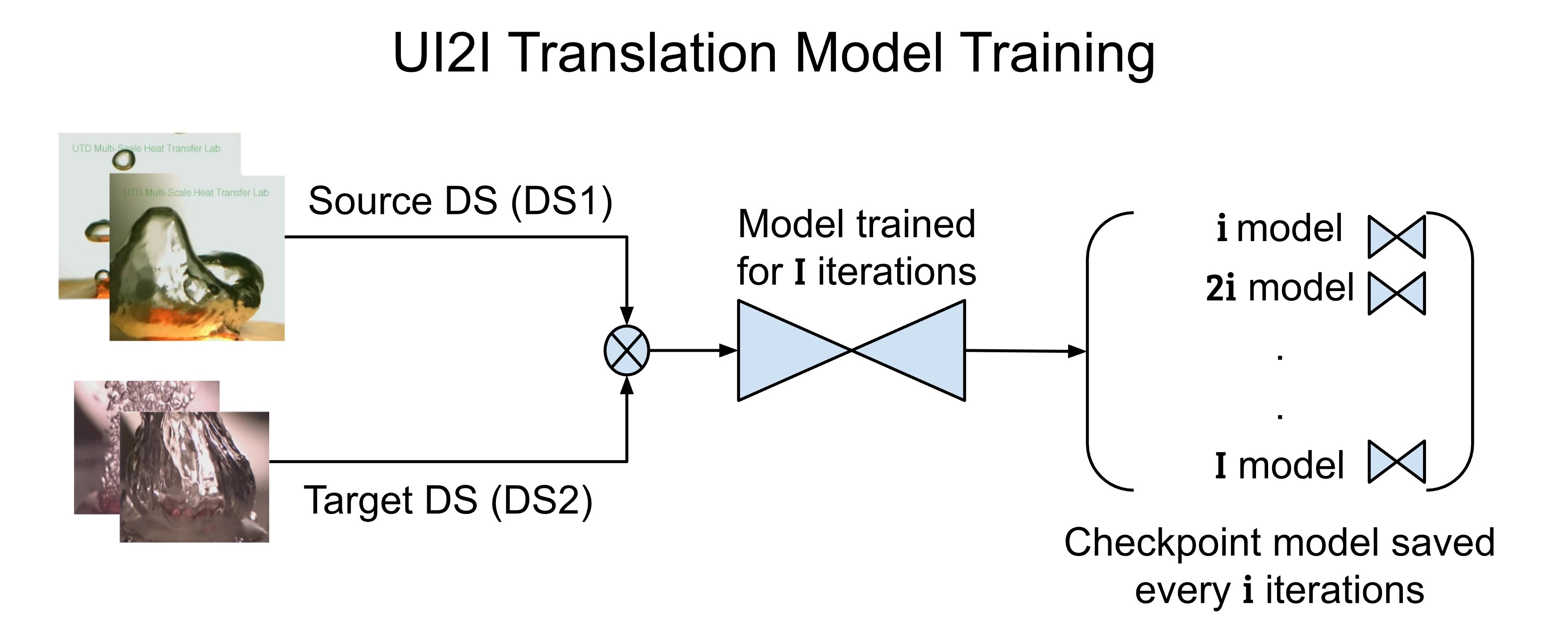}
    \caption{UI2I training process}
    \label{fig:UI2I-training}
\end{figure}

Figure \ref{fig:UI2I-training} summarizes the UI2I translation training process. In order for enabling the source classification model to correctly classify images from the target dataset; which is an unlabeled dataset coming from a different domain that the source classification model has not seen before, an unsupervised Image-to-Image (UI2I) translation Generative Adversarial network (GAN) was employed to translate the images in the target dataset from their domain to the source domain, so that they look like images familiar to the source classification model. For the purpose of demonstrating the methodology Fixed-Point GAN (FP-GAN) \cite{Siddiquee_2019_ICCV} was employed as the UI2I translation model. FP-GAN was designed to support domain adaptation by identifying a minimal subset of pixels for domain translation and has shown superiority over other models in domain translation tasks. This being said, the methodology is agnostic to the type of UI2I translation model used. The UI2I translation model is trained for $I$ number of iterations and a model is saved every $i$ iterations during training, making a total of $I/i$ checkpoint models saved. Both $I$ and $i$ are hyper-parameters that need to be tuned. 

\subsection{Cross-Domain Image Translation Process}
\label{image_translation}

\begin{figure}[H]
    \centering
    \includegraphics[width = \columnwidth, frame]{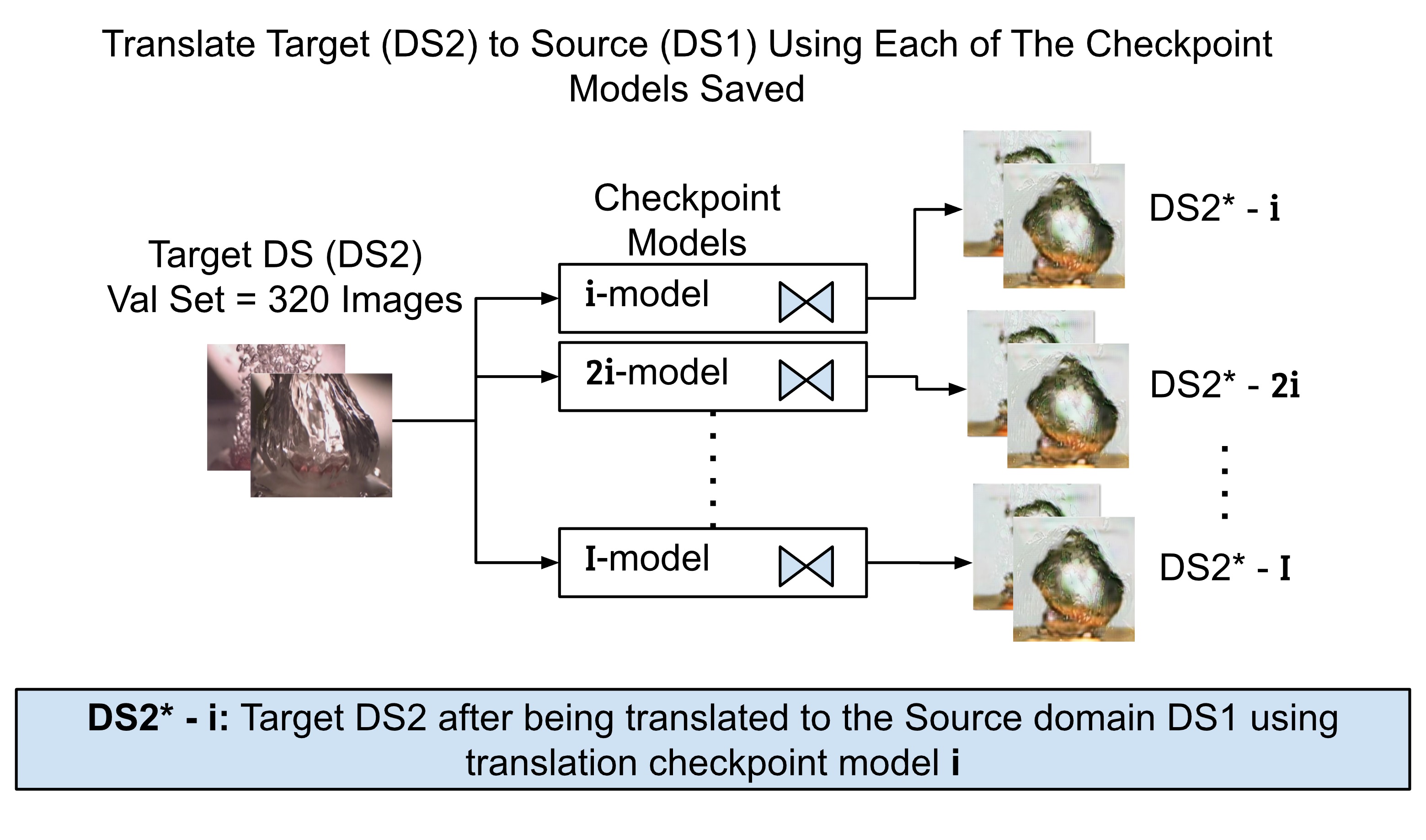}
    \caption{Image translation process}
    \label{fig:Translation-process}
\end{figure}

Once the UI2I translation model training is complete, the framework uses each of the saved i/I checkpoint models to translate the target validation set from the target domain to the source domain as shown in Figure \ref{fig:Translation-process}. This is done in order to evaluate which model is the best one to be used in deployment.

\subsection{Pseudo Supervised Metrics}

\label{Pseudo_eval_sec}

\begin{figure}[H]
    \centering
    \includegraphics[width = \columnwidth, frame]{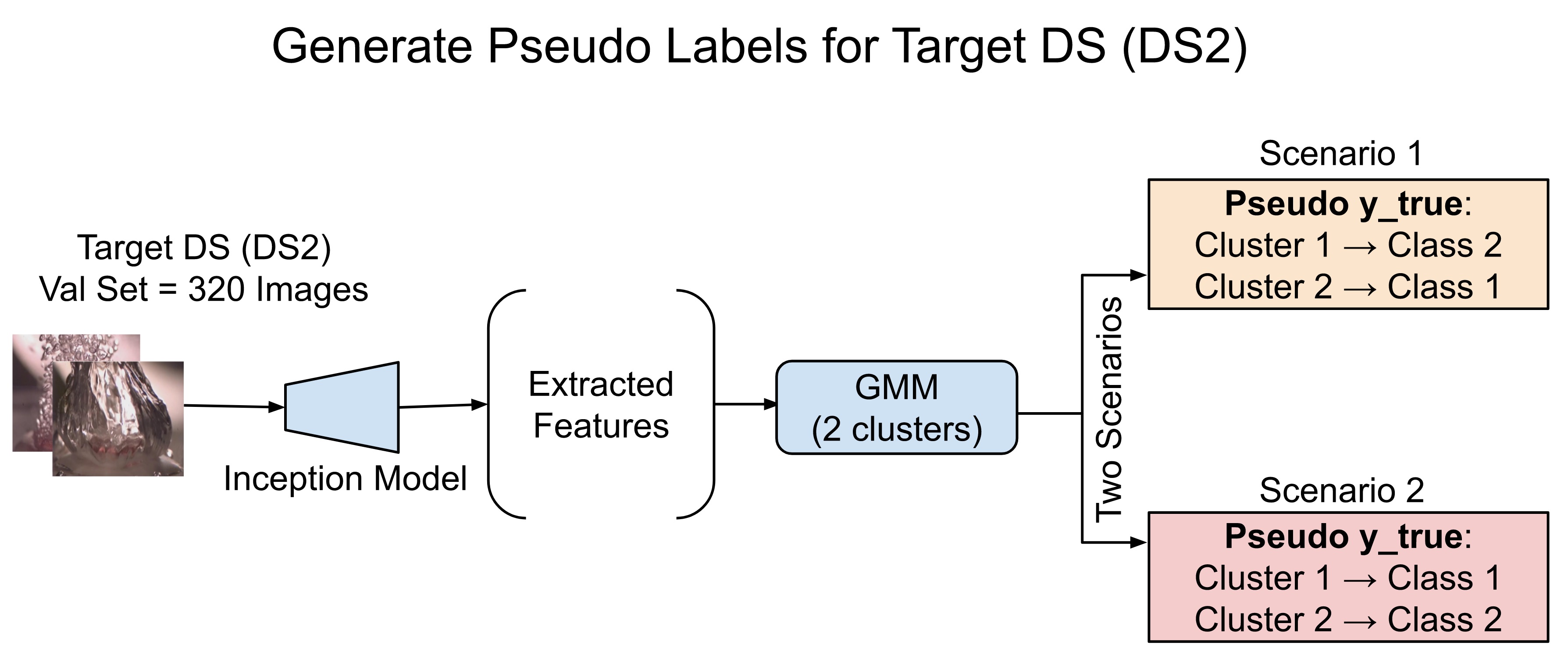}
    \caption{Pseudo labels generation}
    \label{fig:Pseudo Labels}
\end{figure}

The problem now is knowing which of these I/i UI2I translation models to use for model deployment. Since the target dataset is unlabeled, supervised metrics cannot be used to evaluate the validation set. The go-to metric in I2I translation is either the Inception Score (IS) or the Frechet Inception Distance (FID). The problem with these metrics is that they do not guarantee the best model to be selected. To combat this problem, this novel framework is introduced to generate pseudo-labels for the target dataset before translation, which will eventually allow the use of the traditional supervised metrics to evaluate the best model, as demonstrated in Figure \ref{fig:Pseudo Labels}.  Any of the standard supervised metrics used for classification could be used as a pseudo metric once the pseudo labels were generated, but to demonstrate the methodology, the results are showcased using both the balanced accuracy and the AUC metrics and then compared against the IS, FID, KID, and MMD. The metrics used in our experiments and their mathematical definition are listed in table \ref{Metrics}.

\begin{table}[H]
\scriptsize
\centering
\caption{Metrics Definitions}
\renewcommand{\arraystretch}{1.5}

\begin{tabular}{|l|l|}
\hline
\textbf{Metric Name} & \textbf{Mathematical Formulation} \\ 
\hline 
AUC  & $= \int_{-\infty}^{\infty} \text{TPR}(t) \cdot \text{FPR}'(t) \, dt$ \\ 
\hline
Balanced Accuracy & $= \frac{1}{2} \left( \frac{\text{TP}}{\text{TP}+\text{FN}} + \frac{\text{TN}}{\text{TN}+\text{FP}} \right)$ \\ 
\hline
FID  & $= \|\mu_P - \mu_Q\|^2 + \text{Tr}(C_P + C_Q - 2(C_P \cdot C_Q)^{1/2})$ \\ 
\hline
Inception Score & $= \exp\left(\mathbb{E}_x\left[D_{KL}(P(y|x) || P(y))\right]\right)$ \\
\hline
MMD & $\begin{aligned}
&= \frac{1}{m(m-1)} \sum_{i \neq j}^m k\left(x_i, x_j\right) 
\quad + \frac{1}{n(n-1)} \sum_{i \neq j}^n k\left(y_i, y_j\right) \\
&\quad - \frac{2}{m n} \sum_{i=1}^m \sum_{j=1}^n k\left(x_i, y_j\right)
\end{aligned}$ \\
\hline
KID metric & $= MMD_{poly}\left(f_{\text {real }}, f_{\text {fake }}\right)^2$ \\
\hline
\end{tabular}

\label{Metrics}
\end{table}

The method starts by using the pre-trained inception model to generate features from the validation set of the target dataset prior to translation.  Once the features were extracted, the idea is to use a clustering technique to separate the two classes of data, thus creating pseudo labels which will allow using the traditional supervised metrics to evaluate the models. The clustering technique adopted to perform this task was GMMs. Using GMMs to cluster images and image features has been widely adopted in the literature \citep{bakheet2023content,pu2023deep,hou2014novel,kermani2015automatic} for multiple reasons: 1) GMMs are suitable for clustering tasks where the underlying data distribution is not well-defined or contains multiple subpopulations such as the case with boiling images. 2) GMMs can provide robust clustering results even in data consisting of high-dimensional features such as image features. 3) GMMs allow the incorporation of prior knowledge about the data distribution through the initialization of the expected number of clusters. such as in the case of the CHF detection problem. Thus, after extracting the features, a Gaussian Mixture Model is used with a pre-set number of $N$ clusters, where $N$ is the number of classes known from prior domain knowledge. This will group the images into $N$ clusters which are used as pseudo labels (or pseudo-classes) for the unlabeled data. For the purpose of the experiments conducted in this work, the number of classes set by prior domain knowledge is equal to two ($N = 2$).

\begin{figure}[H]
    \centering
    \includegraphics[width = \columnwidth, frame]{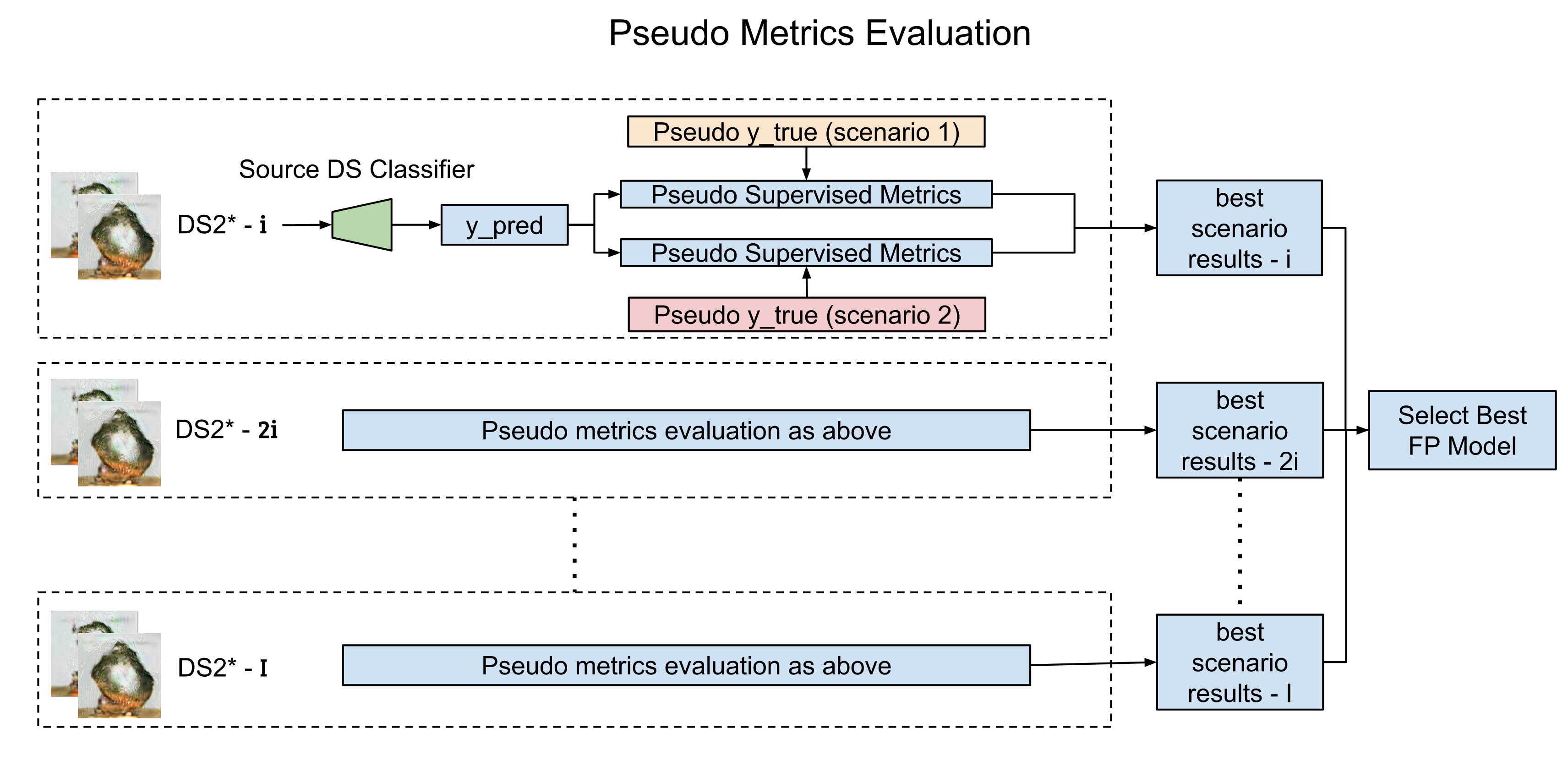}
    \caption{Pseudo metrics evaluation}
    \label{fig:Pseudo metrics}
\end{figure}

Now that the pseudo labels are obtained, they could be used to generate the pseudo-supervised metrics. For each translated validation set of the target dataset, the source classification model is used to generate predictions. The generated predictions are then compared with the pseudo labels using the traditional supervised metrics equations, thus providing "pseudo" supervised metrics. Note that since it is unknown which cluster represents which actual label (class), all possible scenarios are explored. The average of the pseudo metric for all models is then calculated for all scenarios, and the best-scoring scenario is adopted. Finally, the best scoring model from the best scoring scenario is selected for deployment. The pseudo metrics evaluation is described in Figure \ref{fig:Pseudo metrics}.

It is worth mentioning that employing a pre-trained Inception model for feature extraction has the potential of introducing bias towards the ImageNET dataset. However, it's important to note that In contrast to methods like FID which uses summarized statistics of the features in their metric calculation making it more more susceptible to the bias present in the pre-trained model. DIPS does not use neither the features nor their summarized statistics directly in metrics calculations. In DIPS the features obtained from the Inception model are subjected to GMM clustering. The objective here is to group similar features together based on their underlying distribution in the data. GMM clustering operates independently of the original dataset's bias because it seeks to identify patterns and relationships within the feature space, not the dataset from which the features were extracted. Once the clusters are obtained through GMM, the Inception-extracted features are no longer used for metric calculation. Instead, the clusters are relied on to generate pseudo-labels. The subsequent metric calculations are based on these pseudo-labels, not the original features. This step ensures that the bias from the pre-trained Inception model is mitigated and does not directly influence the generated metrics.

\section{Experiment}
\subsection{The Boiling Crisis Detection Problem}

Over the past few decades, the study of heat transfer mechanisms has become a focal topic for researchers around the world. Heat transfer mechanisms are critical in various industrial applications \citep{dirker2019thermal,birbarah2020water,el2012immersion,kandlikar2014review,fenech2013heat}. One of the widely implemented heat transfer mechanisms is boiling heat transfer, a mechanism that utilizes the latent heat of the working fluid to dissipate a large amount of heat with minimal temperature increase \citep{rassoulinejad-mousavi2021a}. Despite its wide application and the amount of effort spent studying boiling heat transfer, this mechanism comes with a dangerous drawback known as the boiling crisis. The boiling crisis is the phenomenon where the heat flux of boiling reaches a critical bound known as the Critical Heat Flux (CHF), after which the heating surface will be covered by a blanket of continuous vapor layer that adversely affects heat dissipation by depreciating the heat transfer coefficient  \citep{al2023framework}. This is critically dangerous because the improper heat dissipation will lead to a quick temperature upraise on the heater surface beyond its capability and eventually cause it to break down. Many efforts were dedicated to investigating the applicability of machine learning algorithms in CHF detection using a variety of techniques and data types. Whether it was acoustic emissions \citep{sinha2021a}, optical images \citep{rokoni2022a}, thermographs \citep{ravichandran2021a} or whether it was using a variety of supervised learning algorithms, including support vector machine \citep{hobold2018a}, multilayer perceptron (MLP) neural networks  \citep{hobold2018a}, transfer learning \citep{rassoulinejad-mousavi2021a}, and most recently researchers \citep{al2023framework} started using frameworks supported by UI2I translation models to solve the cross-domain classification problem in boiling crisis detection such as the example used in this work to showcase our methodology.

\subsection{Dataset}

In this work, two different pool boiling experimental image datasets (DS-1 and DS-2) were prepared, where both DS-1 and DS-2 were generated using publicly available videos \cite{you-a,minseok2014a}. Specifically, the video from which DS-1 was prepared shows a pool boiling experiment performed using a square heater made of high-temperature, thermally-conductive microporous coated copper where the surface was fabricated by sintering copper powder. The square heater had a surface area of $\approx$ $100$ $mm^2$ and the working fluid used was water. All experiments were performed at a steady-state under an ambient pressure of 1 atm. A T-type thermocouple was used for temperature measurements. The resolution of the video frames was $512$ x $480$  pixels. The YouTube video from which DS-2 was prepared shows a pool boiling experiment performed using a circular heater made of microporous-coated copper where the surface was fabricated by sintering copper powder. The circular heater had a diameter of $\approx$ $16$ $mm$ and the working fluid used was DI water. All experiments were performed at a steady state under an ambient pressure of 50 kPa. A T-type thermocouple was used for temperature measurements. The resolution of the video frames was $1280$ x $720$ pixels.

Images for DS-1 and DS-2 were prepared by downloading the videos from YouTube and extracting individual frames using a MATLAB code via the VideoReader and imwrite functions. Recognizing duplicate frames extracted from the YouTube videos, quality control was conducted to remove the repeated images by calculating the relative difference using the Structural Similarity Index (SSIM) value \cite{gao2020a} between two consecutive images where images with a relative difference less than 0.03\% were removed. This pre-processing is important to ensure DL models were not biased by identical image frames. 

\begin{figure}[H]
\captionsetup{justification=centering}
\centerline{\includegraphics[width = \columnwidth]{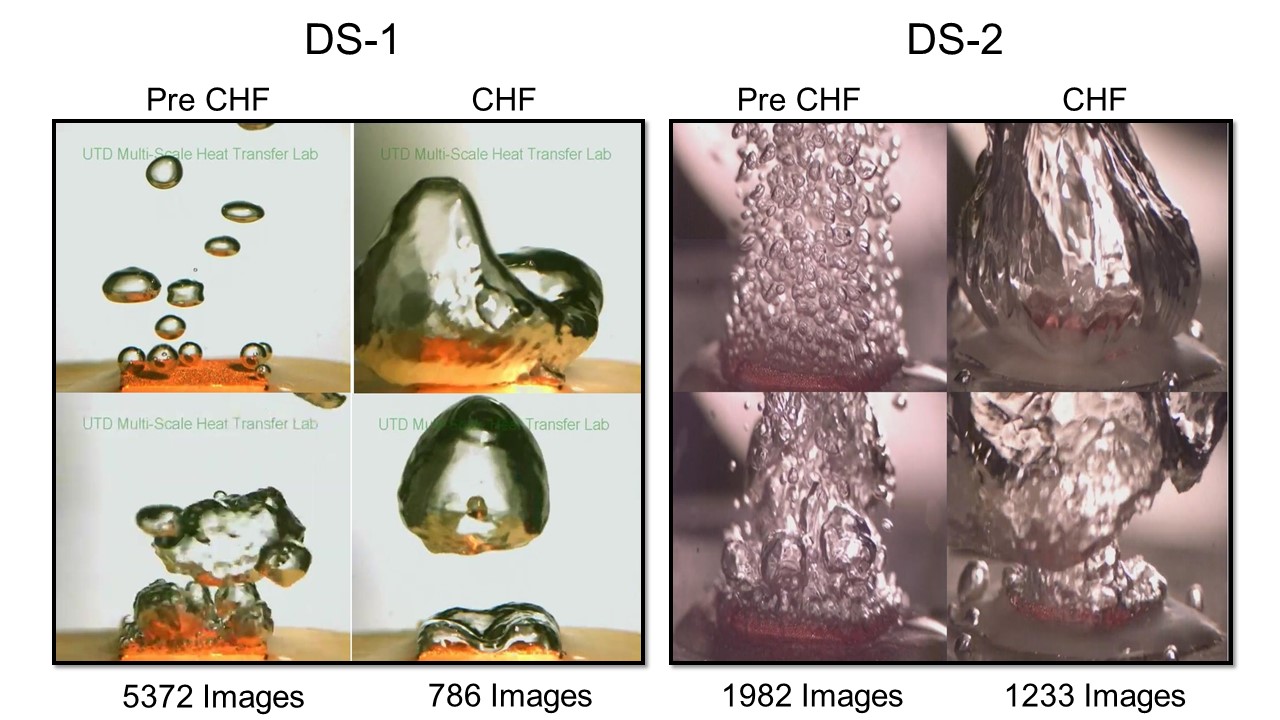}}
\caption{Representative images of bubble dynamics from source videos.}
\label{fig3}
\end{figure}

The images were categorized into two boiling regimes: (1) The critical heat flux regime (CHF), where a significant drop in the heat transfer coefficient is observed due to a continuous vapor layer blanketing the heater surface and (2) pre-CHF regime, where optimal heat transfer coefficient is obtained and  only discrete bubbles or frequent bubble coalescence is observed before departure. Originally, DS-1 had a total of 6158 images (786 CHF versus 5372 pre-CHF) and DS-2 had a total of 3215 (1233 CHF versus 1982 pre-CHF). As seen, both data sets were unbalanced. In each of the two experiments, only the training data for the source dataset was balanced before use. The target dataset was not balanced since the objective of this study is to introduce a framework that utilizes unsupervised learning, that is, the labeling information of the target datasets are assumed to be unavailable; Thus, impossible to balance using traditional oversampling or undersampling techniques. Table\ref{table1} shows the original number of images in each regime for each dataset and Fig\ref{fig3} shows a visual representation of the images for each dataset. The pixel intensity values in each image were normalized to fit in the range [0,1] to ensure uniformity over multiple datasets during deep learning training.

% \begin{figure}[!t] ----> uncomment this when doing two  column paper format

\begin{table}[H]
\captionsetup{justification=centering}
\caption{Datasets Summary}
\label{table1}
\centering
\scriptsize  % --> You may need to remove this line 
\resizebox{\columnwidth}{!}{%
%\begin{tabular}{lcccc}
\begin{tabular}{>{\centering\arraybackslash}m{1cm}>{\centering\arraybackslash}m{2cm}>{\centering\arraybackslash}m{2cm}}
\toprule

\textbf{DS} & \textbf{Pre-CHF} & \textbf{CHF} \\
\midrule
DS-1 & 5372 & 786  \\
DS-2 & 1982 & 1233   \\ 
\bottomrule             
\end{tabular}
}
\end{table}

\subsection{Framework Pipeline}

As depicted in Figure \ref{fig:framework}, the proposed framework consists of four main parts. 

\subsubsection{Source Classification Model Training}
\label{CNN_training_model}

\begin{figure}[H]
\captionsetup{justification=centering}
% \centerline{\includegraphics[width = \columnwidth]{CNN arch_v4.jpg}}
\centerline{\includegraphics[scale= 0.22,fbox]{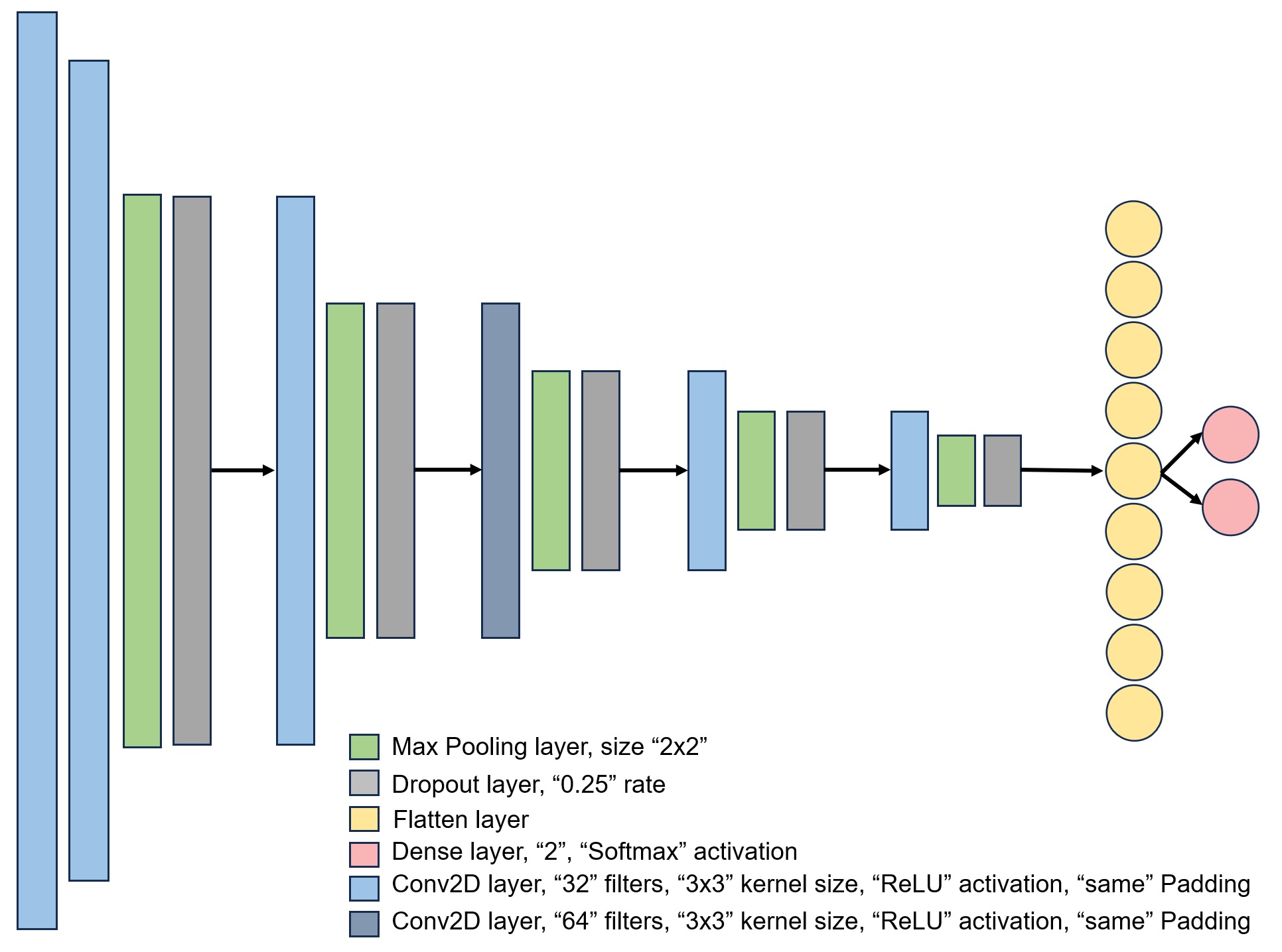}}
\caption{Architecture for the source classification model for the $DS1 \rightarrow{} DS2$ experiment.}
\label{CNN_DS1_2_DS2}
\end{figure}

To diversify the training, different architectures for the classification model were used in each experiment. For the $DS1 \rightarrow{} DS2$ experiment, the same architecture in  \cite{al2023framework} was employed to train the model. The architecture for that model is summarized in figure \ref{CNN_DS1_2_DS2}. For the $DS2 \rightarrow{} DS1$ experiment, the ResNet50 model architecture \cite{He2015} was employed. In both experiments, the data was split into three subsets: a training set (80\%), a validation set (10\%), and a testing set (10\%). The models were trained for a total of 100 epochs and the best model was saved and used in our pipeline. 

\subsubsection{UI2I translation Model Training}

For the UI2I translation, the FP-GAN model was employed. The model was trained using the same architectures for both the generator and the discriminator as implemented by the authors \cite{Siddiquee_2019_ICCV}. Table \ref{exp_settings} summarizes the hyperparameters chosen for training the model and Table \ref{loss_functions} summarizes the loss functions used in the model.

\begin{table}[H]
\centering
\caption{Experiment Settings}

\begin{tabular}{|c|c|}
\hline
\textbf{Parameter} & \textbf{Value} \\ \hline
Image size & 256 x 256 \\ \hline
$C_{dim}$ & $1$ \\ \hline
Batch size & $8$ \\ \hline
Number of workers & $4$ \\ \hline
$\lambda_{id}$ & $0.1$ \\ \hline
Number of iterations ($I$) & $300,000$ \\ \hline
Checkpoint saving frequency ($i$) & $10,000$ \\ \hline
Learning rate for $G$ and $D$ & $0.0001$ \\ \hline
Number of $D$ updates per each G update & $5$ \\ \hline
$\beta_1$ for Adam optimizer & $0.5$ \\ \hline
$\beta_2$ for Adam optimizer & $0.999$ \\ \hline
Data augmentation used & Random horizontal flips \\ \hline
\end{tabular}

\label{exp_settings}
\end{table}

\begin{table}[H]
\centering
\caption{Summary of the loss functions used in FP-GAN.}

\begin{tabular}{|l|l|l|}
\hline
\textbf{Equation} &  \textbf{Loss} &  \textbf{Definition} \\ \hline
Eq. 1 & $L_{adv}   $&$ =\sum_{c \in \{cx, cy\}} E_{x,c} [\log (1 - D_{r/f} (G(x, c)))] + E_x [\log D_{r/f} (x)]$ \\ \hline
Eq. 2 &  $L_{domain}  $&$ = E_{x,cx} [\log D_{domain} (cx | x)]$ \\ \hline
Eq. 3 &  $L_{f_{domain}}  $&$ = \sum_{c \in \{cx, cy\}} E_{x,c} [-\log D_{domain} (c | G(x, c))]$ \\ \hline
Eq. 4 &  $L_{cyc}  $&$ = \sum_{c \in \{cx, cy\}} E_{x,cx} [\|G (G(x, c), cx) - x\|_1]$ \\ \hline
Eq. 5 &  $L_{id}  $&$ = E_{x,c} [\|G(x, c) - x\|_1]$ if $c = cx$; $0$ otherwise \\ \hline
Eq. 6 &  $L_D  $&$ = -L_{adv} + \lambda_{domain} L_{r_{domain}}$ \\ \hline
Eq. 7 &  $L_G  $&$ = L_{adv} + \lambda_{domain} L_{f_{domain}} + \lambda_{cyc} L_{cyc} + \lambda_{id} L_{id}$ \\ \hline
\end{tabular}

\label{loss_functions}
\end{table}

\subsubsection{Pseudo Labels Generation}

The pseudo labels generation process utilizes the InceptionV3 model which was pre-trained on the ImageNET dataset to extract the features from the target DS images after being resized to the dimensions expected by the InceptionV3 model (299x299). Specifically, the features were extracted using the "conv2d\_93" intermediate layer of the model. The extracted features were then clustered into two clusters using the GMM provided by the Scikit-learn library. Other than the number of clusters, the default settings set by the library were used for GMM. Two possible scenarios are then formulated for the true labels (pseudo $y_{true}$): a) cluster 1 = "Pre CHF" and cluster 2 = "CHF" and b) cluster 1 = "CHF" and cluster 2 = "Pre CHF". The two scenarios are then used in the final step. 

\subsubsection{Pseudo Metrics Evaluation}

The labels for each of the translated datasets from the UI2I translation step are predicted ($y_{pred}$ using the same classification model from section \ref{CNN_training_model}. The "balanced accuracy score" and the "roc auc score" from the standard supervised metrics provided by Scikit-learn were used to evaluate $y_{pred}$ against the pseudo $y_{true}$ generated from each scenario. The average of the pseudo metric for all 30 models is then calculated for both scenarios, and the best-scoring scenario is adopted. Finally, the best scoring model from the best scoring scenario is selected for deployment.

\section{Results and Discussion}

Once the pseudo metrics were generated, the models generated by the FP-GAN training process can now be ranked according to these metrics and the best model can now be selected. Any of the regular metrics used for classification could be used as a pseudo metric once the pseudo labels were generated, but to demonstrate the methodology, the results are showcased using both the balanced accuracy and the AUC metrics.

% In order to showcase the efficacy of the framework, three questions need to be answered: 1) Does the proposed metric outperform the state-of-the-art GAN evaluation metrics in the field (FID, KID, IS, MMD)?, 2) How does it compare against the best possible scenario, that being the true supervised metrics calculated using the true labels, and 3) How does the proposed metric correlate with the actual true metric in comparison to the state-of-the-art?. In other words, had the actual annotations been accessible to enable the use of the standard supervised metrics, how will the proposed metric compare, while keeping in mind that the proposed framework has no access to these annotations in deployment?

To demonstrate the effectiveness of the framework, three key aspects must be addressed. Firstly, it must be assessed whether the proposed metric performs better than the current state-of-the-art GAN evaluation metrics in the field, including FID, KID, IS, and MMD. Secondly, it is essential to compare the proposed metric with the ideal scenario, where true supervised metrics are calculated using the actual labels. Finally, it should be examined how the proposed metric correlates with the actual true metric when compared to how the existing state-of-the-art metrics correlate with the actual true metric. This section addresses these aspects in two ways. Firstly, each of the saved checkpoint models is evaluated using the truly-supervised metric, pseudo-supervised metric, FID, KID, IS, MMD (linear Kernel), and MMD (Gaussian kernel). The models are then ranked according to each of the competing metrics and are plotted against the true actual supervised score of the ranked checkpoint models. In this comparison, each metric is judged by how close it is to mimicking the ranking behavior of the truly supervised metrics that had access to the annotations; thus allowing for an objective evaluation of both the proposed metric and the state-of-the-art metrics. The best metric is the one that could mimic the monotonically decreasing behavior of the true ranking line. 
Secondly, the true metric value is plotted against all the competing metrics and multiple linear and non-linear correlation values were generated for each plot, including R-squared, Pearson correlation, Spearman correlation, and Kendall rank correlation. The rationale here is to showcase the degree of correlation between the true supervised metric and all the competing metrics. The higher the correlation the better the metric is. The results in this section show that the proposed method not only outperforms the state-of-the-art but also heavily resembles the true ranking behavior of the true supervised metrics and is highly correlated with the true supervised metrics.

\begin{figure}[H]
\captionsetup{justification=centering}
% \centerline{\includegraphics[width = \columnwidth]{ds1_ds2BalancedAccuracy_Ranking_plots.jpg}}
\centerline{\includegraphics[scale = 0.5]{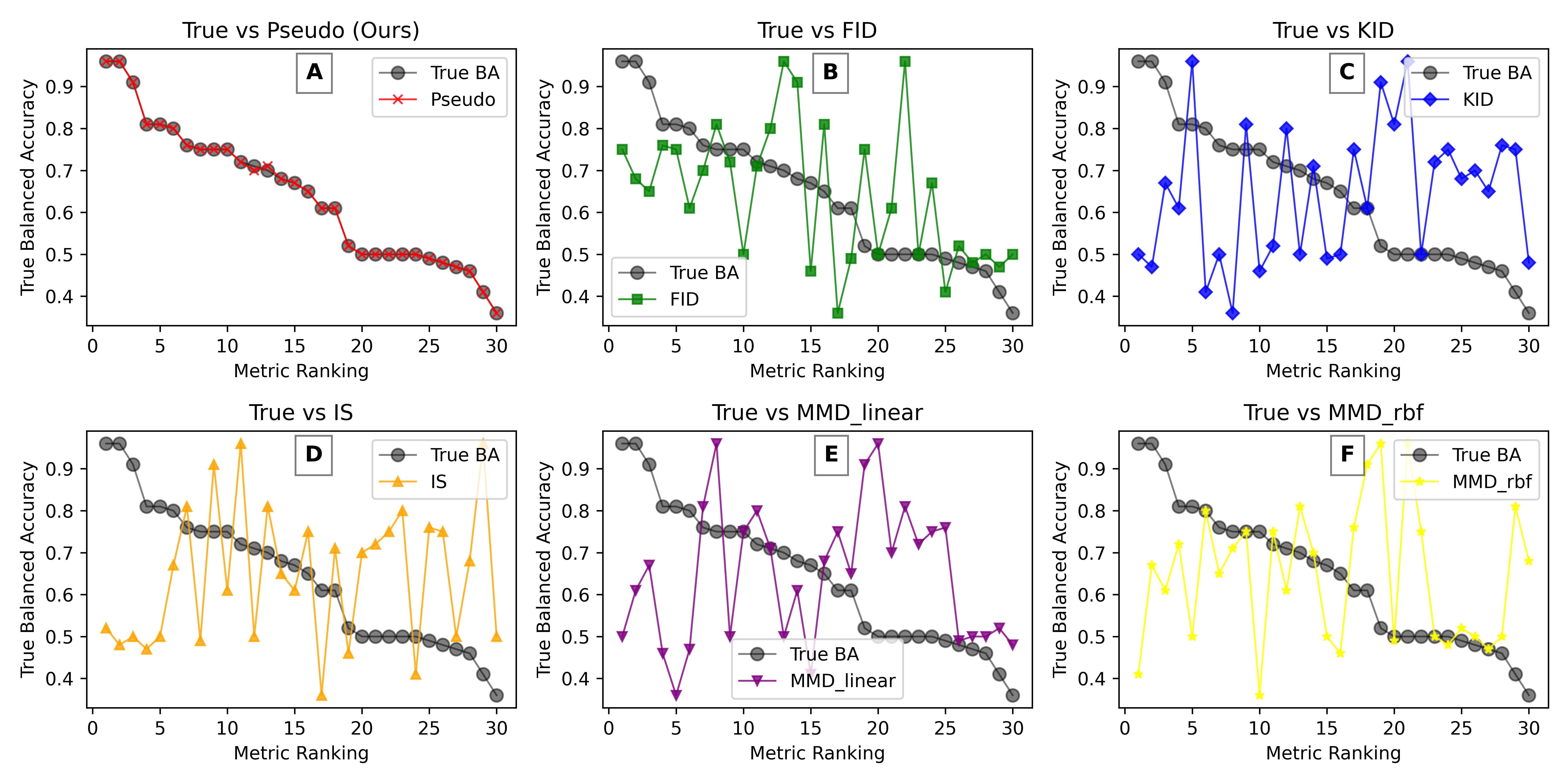}}
\caption{Models ranking results vs the true balanced accuracy for A) pseudo-balanced accuracy (ours), B) FID, C) KID, D) IS, E) MMD (linear kernel), and F) MMD (gaussian kernel) for the $DS1 \rightarrow{} DS2$ Experiment.}
\label{DS1 vs DS2 BA}
\end{figure}

\begin{figure}[H]
\captionsetup{justification=centering}
% \centerline{\includegraphics[width = \columnwidth]{ds1_ds2AUC_Ranking_plots.jpg}}
\centerline{\includegraphics[scale = 0.5]{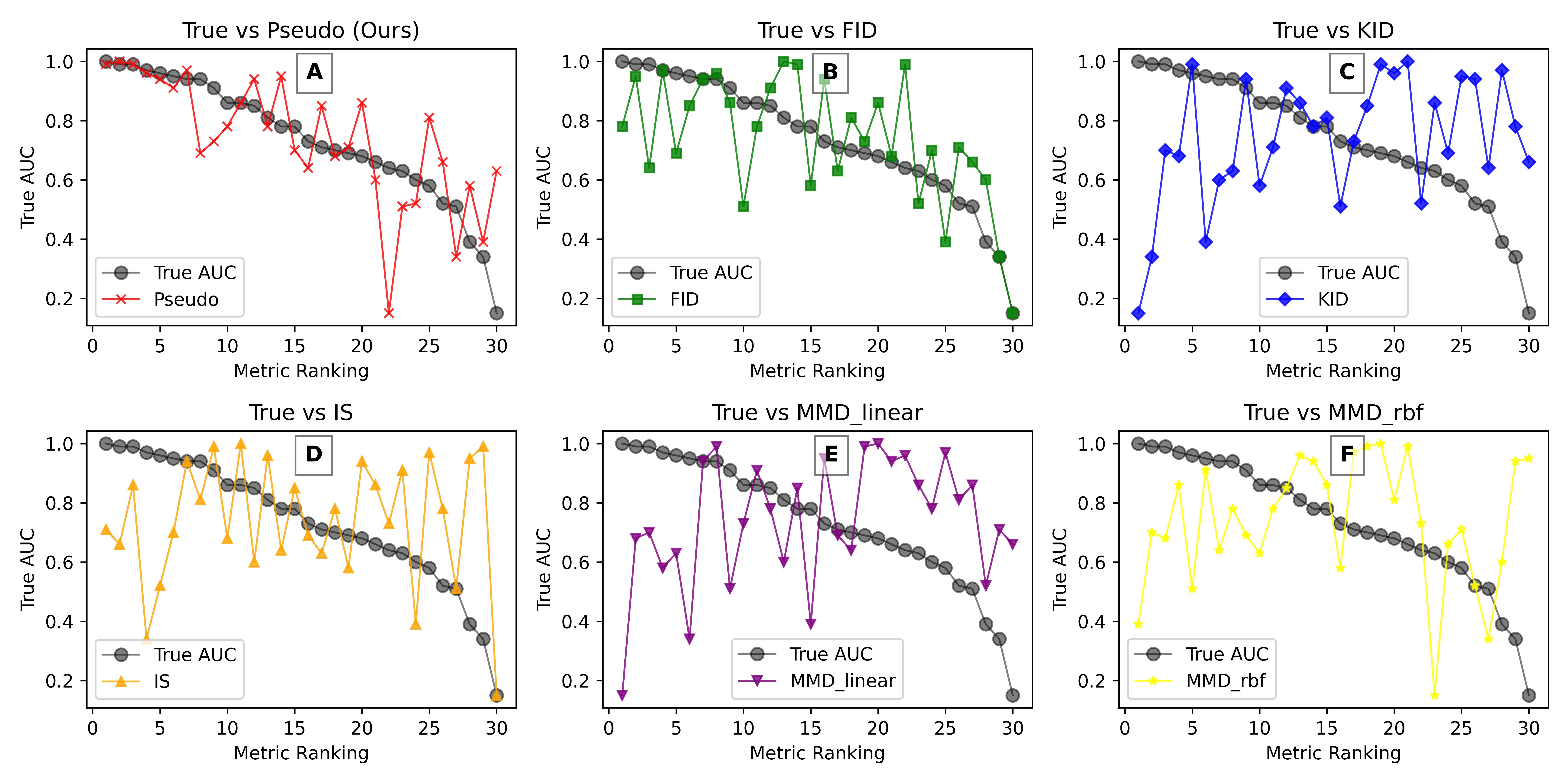}}
\caption{Models ranking results vs the true AUC for A) pseudo-AUC (ours), B) FID, C) KID, D) IS, E) MMD (linear kernel), and F) MMD (gaussian kernel) for the $DS1 \rightarrow{} DS2$ Experiment.}
\label{DS1 vs DS2 AUC}
\end{figure}

\begin{figure}[H]
\captionsetup{justification=centering}
% \centerline{\includegraphics[width = \columnwidth]{ds2_ds1BalancedAccuracy_Ranking_plots.jpg}}
\centerline{\includegraphics[scale = 0.5]{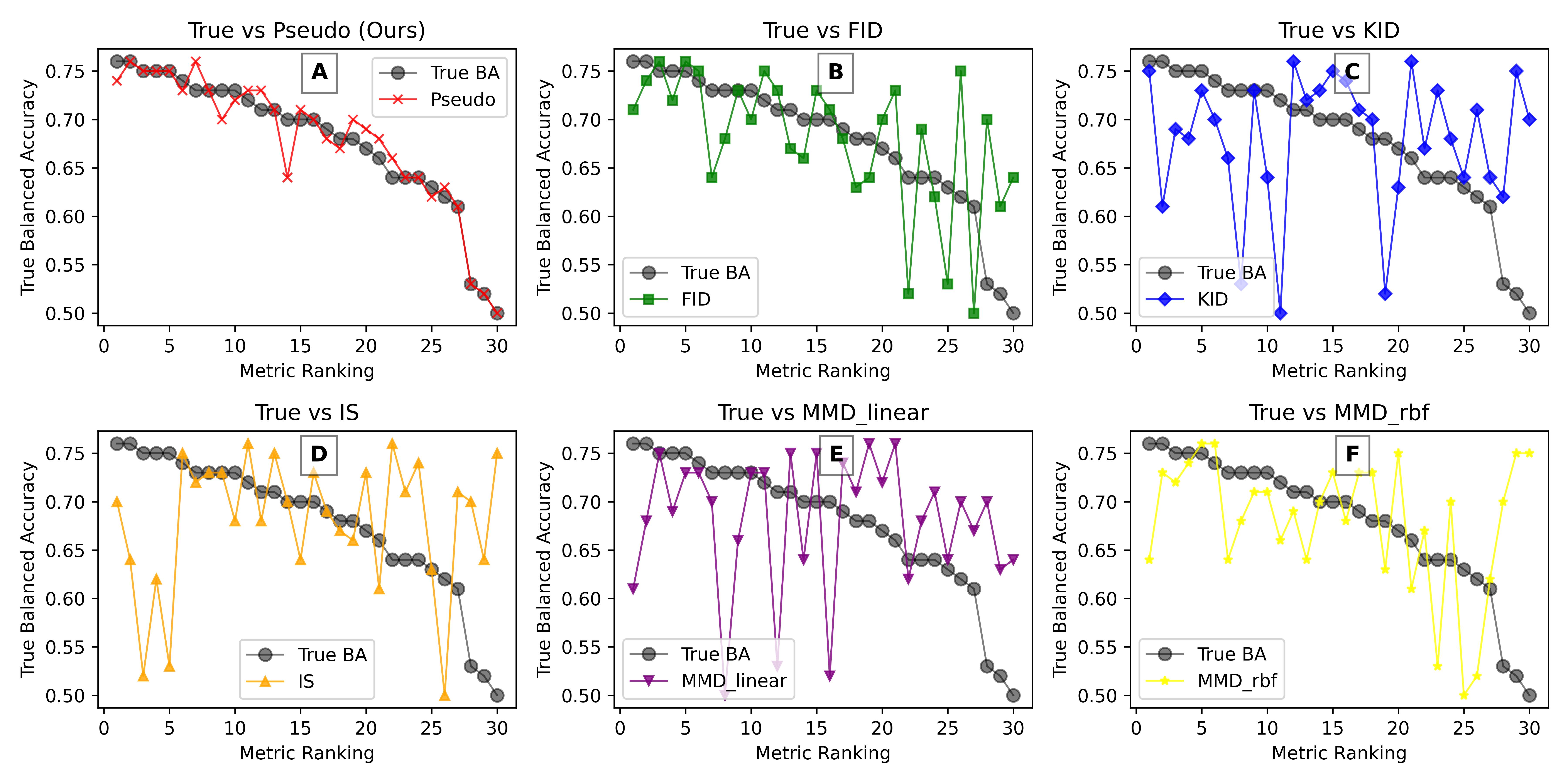}}
\caption{Models ranking results vs the true balanced accuracy for A) pseudo-balanced accuracy (ours), B) FID, C) KID, D) IS, E) MMD (linear kernel), and F) MMD (gaussian kernel) for the $DS2 \rightarrow{} DS1$ Experiment.}
\label{DS2 vs DS1 BA}
\end{figure}

\begin{figure}[H]
\captionsetup{justification=centering}
% \centerline{\includegraphics[width = \columnwidth]{ds2_ds1AUC_Ranking_plots.jpg}}
\centerline{\includegraphics[scale =0.5]{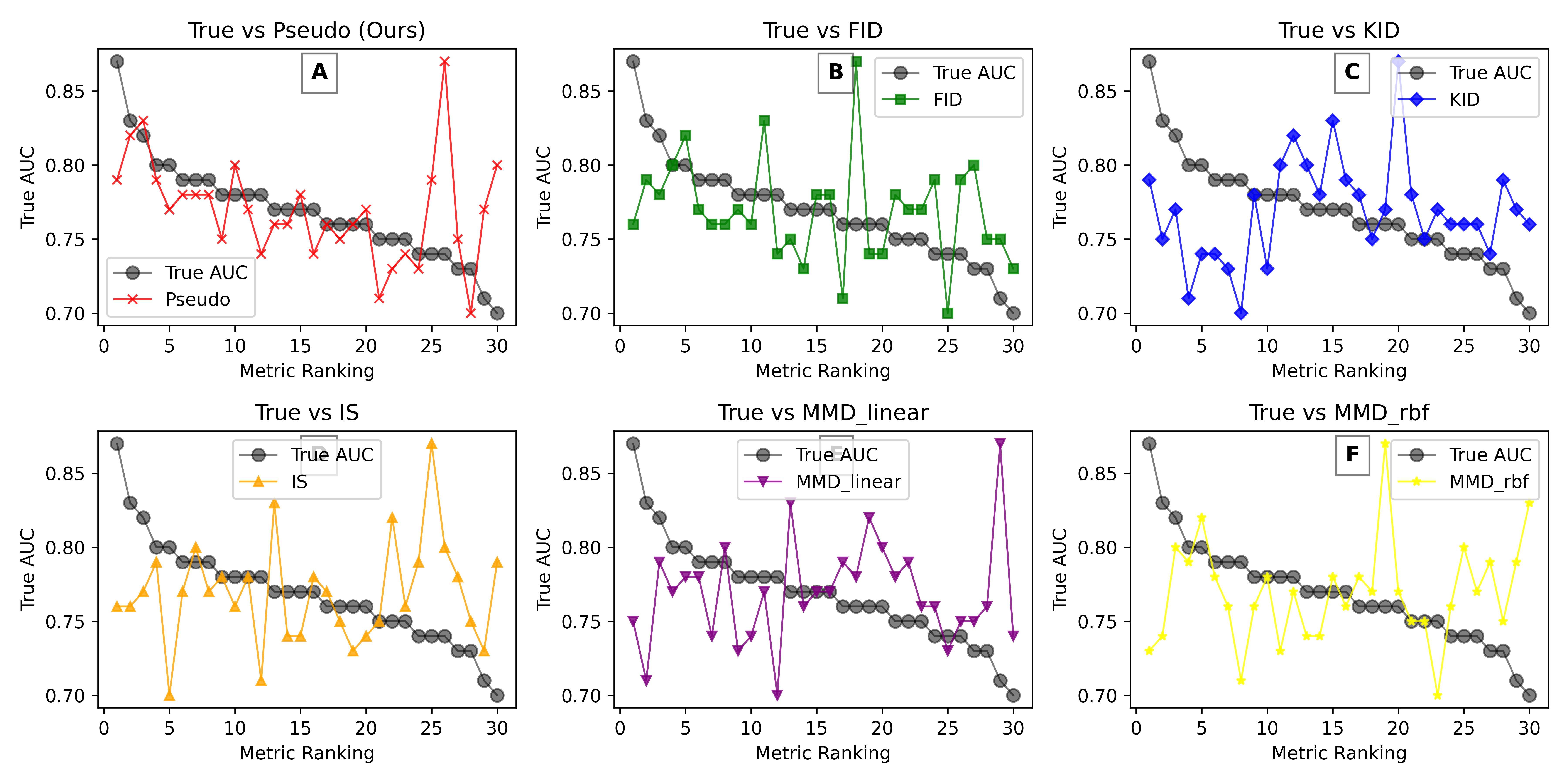}}
\caption{Models ranking results vs the true AUC for A) pseudo-AUC (ours), B) FID, C) KID, D) IS, E) MMD (linear kernel), and F) MMD (gaussian kernel) for the $DS2 \rightarrow{} DS1$ Experiment.}
\label{DS2 vs DS1 AUC}
\end{figure}

Figure \ref{DS1 vs DS2 BA} shows the models' ranking results based on the true-supervised balanced accuracy for all competing metrics in the experiments ran with DS1 as a source Dataset ($DS1  \rightarrow{} DS2$). Figure \ref{DS1 vs DS2 AUC} shows the models' ranking results based on the true-supervised AUC for the same experiment. Figures \ref{DS2 vs DS1 BA} and \ref{DS2 vs DS1 AUC} show the same plots but for the ($DS2  \rightarrow{} DS1$) experiment where DS2 was used as the source dataset. The $x-axis$ represents the model ranking according to that metric from best to worst, while the $y-axis$ represents the real metric value of the ranked model. The ranking curve for the actual ranking will show a monotonically decreasing behavior. In all the mentioned plots, the true actual ranking is compared against A) the pseudo-balanced accuracy (ours), B) the FID, C) the KID, D) the IS, E) the MMD (linear kernel), and F) the MMD (gaussian kernel).

% \begin{figure}[h!]
% \begin{figure}[H]
% \captionsetup{justification=centering}
% \centerline{\includegraphics[width = \columnwidth]{DS1 vs DS2_all.jpg}}
% \caption{Ranking results: A) True balanced accuracy for ranked models when using DS1 as source, B) True AUC for ranked models when using DS1 as source, C) True Balanced Accuracy for ranked models when using DS2 as source, D) True AUC for ranked models when using DS2 as source.}
% \label{DS1 vs DS2_all}
% \end{figure}

As seen from the figures, the pseudo metric ranking is consistent for both the balanced accuracy and the AUC metrics while all the other metrics show very inconsistent and random behavior and demonstrating that indeed none of these metrics are suitable for cross-domain classification. The robustness of the pseudo metrics is more evident when DS2 was used as a source dataset, even when the pseudo metrics deviate from the original ranking, they still pick a model that is close in terms of the actual metric value and returns to mimic the actual ranking behavior. Table \ref{tab:metrics} shows the true balanced accuracy and AUC results of the picked model using each competing metric for both experiments. Again, the proposed metric demonstrate both consistency and superiority over other metrics. Even when the metric comes second (only by a value of 0.01) as in the case of the balanced accuracy for the $DS2 \rightarrow DS1$ experiment, it still picks a very close model to the actual best. Moreover, it shows how inconsistent the state-of-the-art metrics are in cross-domain classification applications.

\begin{table}[H] 
\centering
\caption{Competing metrics comparison (the best is in bold and the second best is underlined).}
\label{tab:metrics}
\scriptsize

\begin{tabular}{ccccccccc}
\hline
\multirow{2}{*}{Exp} & \multirow{2}{*}{Metric} & \multirow{2}{*}{True} &\multicolumn{6}{c}{Competing Metrics} \\
\cline{4-9}
 &  &  & Pseudo & FID & KID & IS & $MMD_{L}$ & $MMD_{G}$ \\
\hline
\multirow{2}{*}{$DS1 \rightarrow DS2$} & Bal. Acc. & 0.96 & \textbf{0.96} & \underline{0.75} & 0.50 & 0.52 & 0.50 & 0.41 \\
 & AUC & 1.00 & \textbf{0.99} & \underline{0.78} & 0.15 & 0.71 & 0.15 & 0.39 \\
 \hline
\multirow{2}{*}{$DS2 \rightarrow DS1$} & Bal. Acc. & 0.76 & \underline{0.74} & 0.71 & \textbf{0.75} & 0.70 & 0.61 & 0.64 \\
 & AUC & 0.87 & \textbf{0.79} & \underline{0.76} & \textbf{0.79} & \underline{0.76} & 0.75 & 0.73 \\
\hline
\end{tabular}

\end{table}

\begin{figure}[H]
\captionsetup{justification=centering}
% \centerline{\includegraphics[width = \columnwidth]{ds1_ds2BalancedAccuracy_Corr_plots.jpg}}
\centerline{\includegraphics[scale = 0.5]{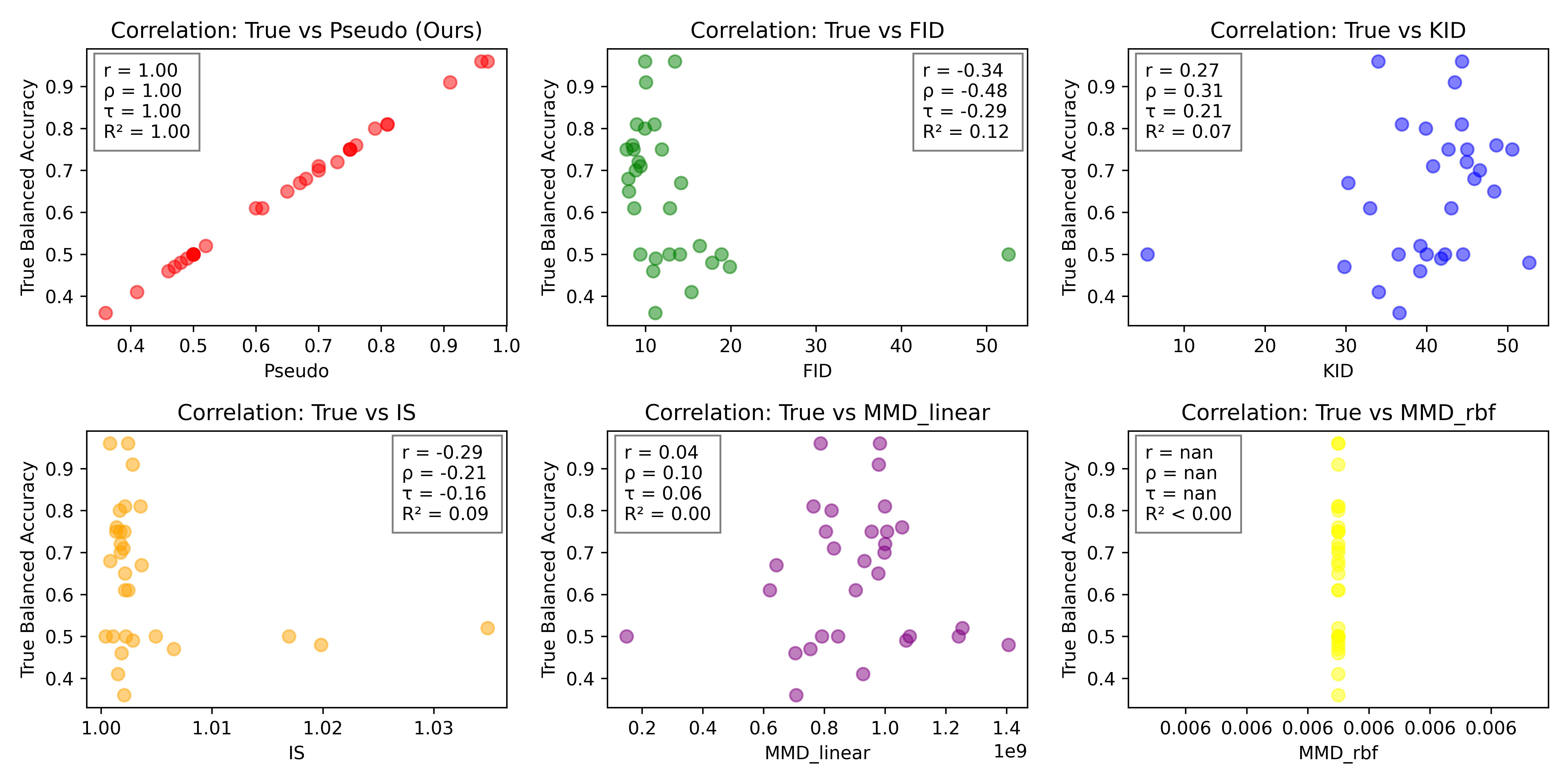}}
\caption{Correlations results of true balanced accuracy vs A) pseudo-balanced accuracy (ours), B) FID, C) KID, D) IS, E) MMD (linear kernel), and F) MMD (gaussian kernel) for the $DS1 \rightarrow{} DS2$ Experiment.}
\label{DS1_DS2 BA corr}
\end{figure}

\begin{figure}[H]
\captionsetup{justification=centering}
% \centerline{\includegraphics[width = \columnwidth]{ds1_ds2AUC_Corr_plots.jpg}}
\centerline{\includegraphics[scale = 0.5]{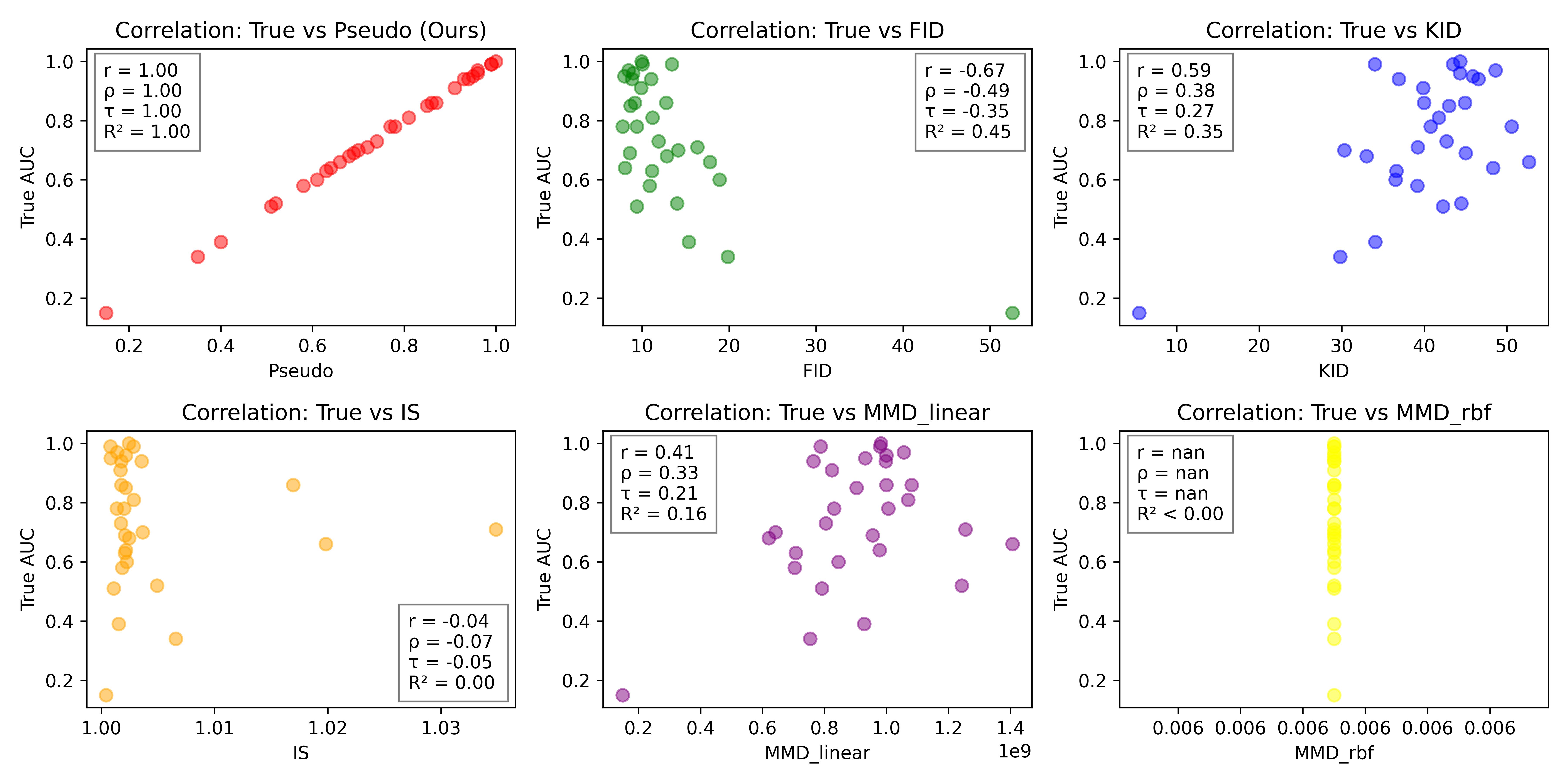}}
\caption{Correlations results of true AUC vs A) pseudo-AUC (ours), B) FID, C) KID, D) IS, E) MMD (linear kernel), and F) MMD (gaussian kernel) for the $DS1 \rightarrow{} DS2$ Experiment. }
\label{DS1_DS2 AUC corr}
\end{figure}

\begin{figure}[H]
\captionsetup{justification=centering}
% \centerline{\includegraphics[width = \columnwidth]{ds2_ds1BalancedAccuracy_Corr_plots.jpg}}
\centerline{\includegraphics[scale = 0.5]{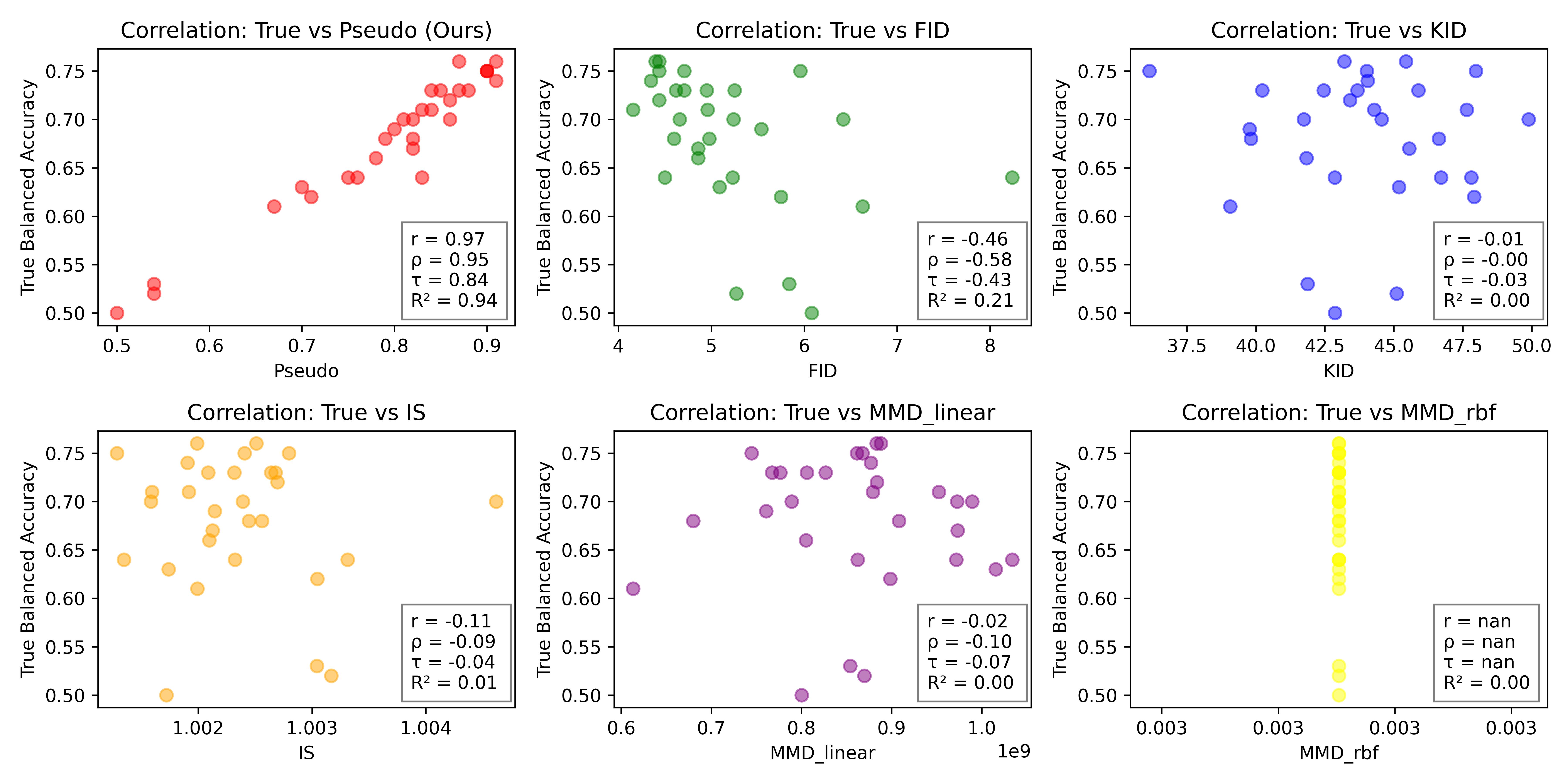}}
\caption{Correlations results of true balanced accuracy vs A) pseudo-balanced accuracy (ours), B) FID, C) KID, D) IS, E) MMD (linear kernel), and F) MMD (gaussian kernel) for the $DS2 \rightarrow{} DS1$ Experiment.}
\label{DS2_DS1 BA corr}
\end{figure}

\begin{figure}[H]
\captionsetup{justification=centering}
% \centerline{\includegraphics[width = \columnwidth]{ds2_ds1AUC_Corr_plots.jpg}}
\centerline{\includegraphics[scale = 0.5]{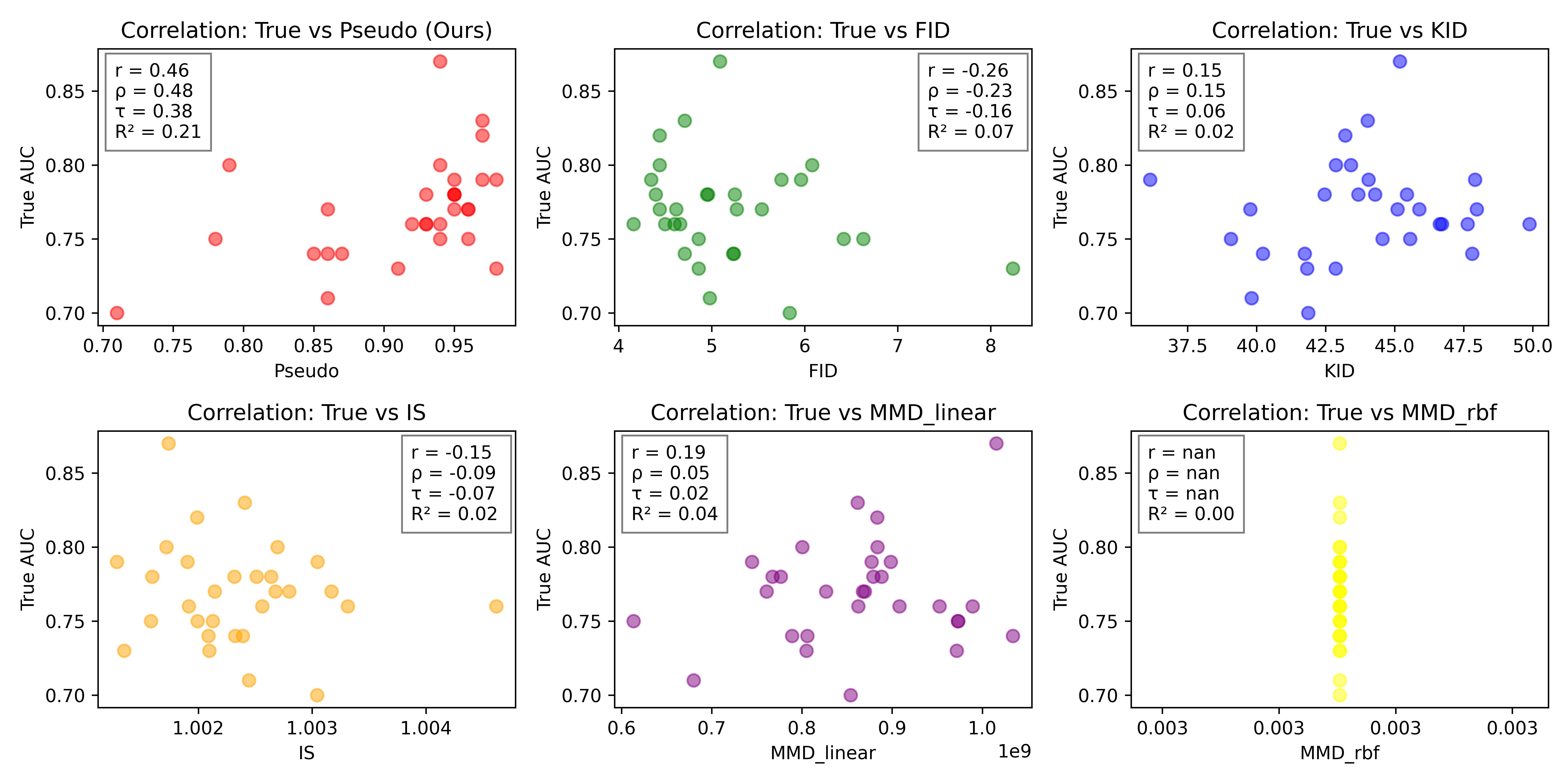}}
\caption{Correlations results of true balanced accuracy vs A) pseudo-AUC (ours), B) FID, C) KID, D) IS, E) MMD (linear kernel), and F) MMD (gaussian kernel) for the $DS2 \rightarrow{} DS1$ Experiment.}
\label{DS2_DS1 AUC corr}
\end{figure}

Figure \ref{DS1_DS2 BA corr} shows the correlation results based on the true-supervised balanced accuracy vs all competing metrics in the experiments ran with DS1 as a source Dataset ($DS1  \rightarrow{} DS2$). Figure \ref{DS1_DS2 AUC corr} shows the same only using true-supervised AUC for the same experiment. Similarly, Figures \ref{DS2_DS1 BA corr} and \ref{DS2_DS1 AUC corr} show the correlations but for the ($DS2  \rightarrow{} DS1$) experiment where DS2 was used as the source dataset. The $x-axis$ represents the metric value while the $y-axis$ represents the real metric value. For each plot, the coefficient of determination ($R^2$), Pearson ($r$), Spearman ($\rho$), and Kendall ($\tau$) correlations were calculated and presented on the plot.

The results from Figures \ref{DS1_DS2 BA corr} and \ref{DS1_DS2 AUC corr} show that for both the pseudo metrics there is an almost perfect correlation with the truly supervised metrics for the $DS1  \rightarrow{} DS2$ experiment. On contrast, all the other competing metrics show weak to none existing correlation with either the AUC or the balanced accuracy.

The results in Figure \ref{DS2_DS1 BA corr} show that again the results for the pseudo metric is strongly correlated with the truely supervised balanced accuracy in the $DS2 \rightarrow{} DS1$ experiment, while again the other competing metrics show weak to none existing correlation with the truly supervised balanced accuracy. Although both the FID and KID showed some improvement in the correlation values, this improvement is not to a level that would qualify them as a suitable candidate for cross-domain classification tasks and their performance remains far worse than that of the pseudo metric. 

The correlation performance of all metrics against the truly supervised AUC for the $DS2  \rightarrow{} DS1$ experiment is shown to be poor in Figure \ref{DS2_DS1 AUC corr}. However, the performance of the pseudo metric is still far better than all the other metrics on all of the correlation measures. One thing to note about this particular experiment is that while we acknowledge the presence of a weak correlation between our pseudo metric and the true metric, it's essential to highlight the uniqueness of the results in this specific experiment. The true metric results for all 30 models demonstrated an exceptionally tight cluster ($\mu = 0.77$ and $std = 0.035$). This consistency suggests that the models' performances were very close to one other according to the true metric. In such scenarios where the true metric values are essentially constant across all models, correlation measures may not be the most appropriate yardstick for assessing the worth of a metric; as in such cases, the performance remains consistently similar irrespective of which checkpoint model is chosen making the choice of the checkpoint model inconsequential in terms of the true metric. This explains why the pseudo metric exhibited an underperformance in this experiment.

The results from the correlation figures demonstrate yet again the superiority of the proposed metric over the state-of-the-art. Moreover, they also indicate that the proposed metrics are explainable since they are highly correlated with the explainable true supervised metrics. Moreover, the results also demonstrate that none of the current state-of-the-art metrics are consistent and they are uncorrelated with the true metrics, making their decision unexplainable and unreliable to assess UI2I translation models for unsupervised cross-domain classification tasks and should not be utilized in such applications.

% The results from the correlation figures demonstrate the superiority of the proposed metric over the state-of-the-art. Moreover, they also indicate that the proposed metrics are explainable since they are highly correlated with the explainable true supervised metrics. One thing to note is that all correlation measures showed high correlation values for the pseudo metrics for all experiments except the pseudo-AUC in the $DS2  \rightarrow{} DS1$ experiment. What is interesting about that experiment is that the results of the real AUC for all 30 FP-GAN models were very close to one another ($\mu = 0.77$ and $std = 0.035$) resulting in a semi-constant line as shown in Figure \ref{DS2 vs DS1 AUC}.

\begin{figure}[H]
\captionsetup{justification=centering}
\centerline{\includegraphics[scale=0.6]{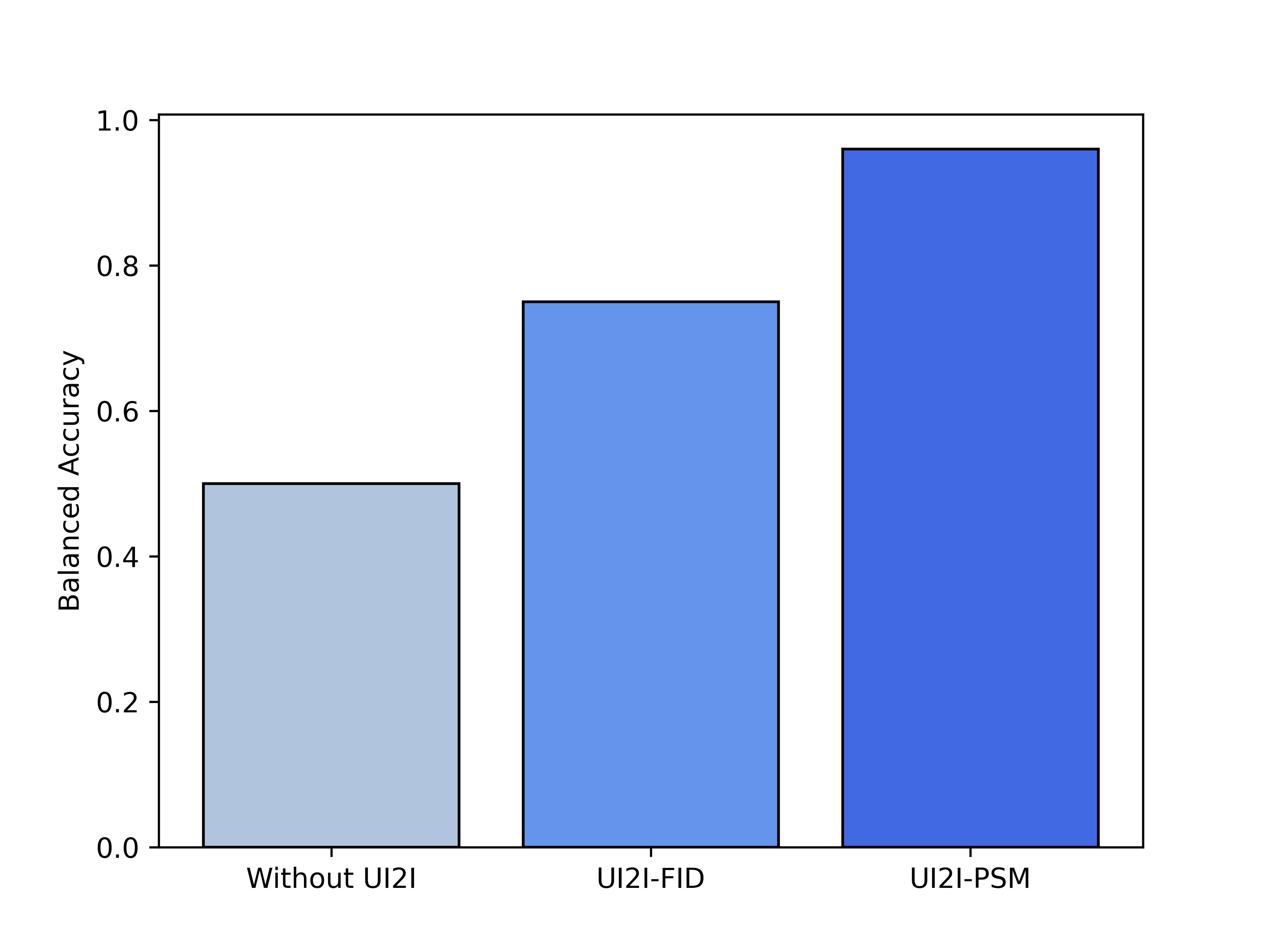}}
\caption{Comparison of classification accuracy of CNN trained on DS-1 for classifying the boiling regime of DS-2 with no translation, UI2I with FID, and UI2I with the developed pseudo supervised metric (PSM).}
\label{bar_plot}
\end{figure}

The pseudo-supervised metric developed in this work will make a major impact on the boiling community. While many research groups have demonstrated visualization-based boiling regime classification or boiling crisis detection using AI models, most of the studies are based on single datasets, leaving model generalizability a major challenge \citep{hobold2018a,ravichandran2019a}. As explored in the authors’ previous work, a well-trained CNN model may only lead to an accuracy of 0.4 – 0.5 when classifying new data sets that are not included in the training \citep{rassoulinejad-mousavi2021a,al2023framework}. This accuracy can be improved using transfer learning, where a small amount of labeled data from the new data set are used to fine-tune pre-trained classifiers \citep{rassoulinejad-mousavi2021a}. UI2I using GAN does not require labeled data from the new data sets but only leads to an accuracy of 0.75 when using state-of-the-art metric (FID) to select the best-performing generator \citep{al2023framework}. The present work using the pseudo-supervised metric can lead to a significantly higher balanced accuracy of 0.96 and thus demonstrates that using UI2I with the pseudo-supervised metric, a pre-trained boiling regime classifier can be adapted to any new boiling data set without additional training or labeled data (see Figure \ref{bar_plot}). 

\section{Conclusion and Future Work}

In this paper, a framework was introduced to evaluate UI2I translation models. The framework was designed specifically to support cross-domain classification applications using pseudo-supervised metrics. To showcase the efficiency of the framework, the boiling crisis detection problem was used as an example. The efficacy of the results was demonstrated by conducting two experiments using two publicly available datasets from different domains.

The proposed methodology was shown to not only outperform the state-of-the-art unsupervised metrics but was also shown to be highly correlated with the true supervised metrics, unlike the state-of-the-art which were poorly correlated with the true supervised metrics and were shown to be inconsistent. Moreover, it was displayed that typical state-of-the-art GAN evaluation metrics which were designed to evaluate models based on their ability to generate images that are both diverse and realistic to the human eye are not suitable to support cross-domain classification tasks as presented in the results section. In almost all comparisons, the ranking provided by the pseudo metric was superior to the state-of-the-art metrics and showed that it can mimic the monotonically decreasing behavior of the true metric; thus providing explainable and consistent results.

Although the proposed metric is showing great potential in evaluating UI2I translation models in cross-domain prediction frameworks, it still has room for improvement and future work. For starters, the proposed metric was only demonstrated to work in a binary classification setting ($N = 2$). An important factor to consider in the case of multi-class classification is that the problem becomes more computationally expensive as the number of clusters $N$ increases. Specifically, in the cluster assignment step of the framework, the number of possible scenarios is expected to increase exponentially. For future work, we plan on addressing both of these issues by incorporating physics-assisted labeling extracted from either a physical source such as acoustic sensors mounted on the apparatus of the experiment, or extracted from the images themselves such as the count and size of the blobs in the image using computer vision segmentation techniques. Another issue is that the framework is dependent on having prior domain knowledge to set up the number of expected clusters $N$ that are later used as pseudo labels. The method falls short in the case where no prior domain knowledge is available about the expected number of labels. Again, physics-assisted labeling extracted using segmentation could provide a solution to this issue. Furthermore, the use of a pre-trained Inception model for feature extraction introduces the potential for bias toward the ImageNET dataset. Although DIPS takes measures to mitigate this bias effect. it's important to acknowledge that there may still be some residual influence from the pre-trained model. Lastly, the method is strictly applicable to cross-domain classification problems and is not suitable to address cross-domain regression problems where the predicted value is continuous rather than discrete. If the data in the source DS is instead labeled with quantifiable heat flux value, then an interesting direction would be to explore the utilization of the temporal factor to relate each frame from the target DS with the source. Researchers are advised to study the mentioned issues and explore the capability of this metric to expand on its potential beyond this work.

% \section{Conclusion}

% In this paper, we introduce a framework to evaluate UI2I translation models that was designed specifically to support cross-domain classification applications using pseudo-supervised metrics. To showcase the efficiency of the framework, we use the boiling crisis detection problem as an example. We demonstrate the consistency of the results by conducting six different experiments using three datasets from different domains.

% We show that our methodology not only outperforms unsupervised metrics such as the FID, but also is highly correlated with the true supervised metrics, unlike the FID which is poorly correlated with the true supervised metrics and inconsistent. Moreover, we show that typical GAN evaluation metrics such as the FID which were designed to evaluate models based on their ability to generate images that are both diverse and realistic to the human eye are not suitable to support cross-domain classification tasks As shown in the results section. We compare our method against the FID metric and the true metrics in cross-domain classification tasks and we show that in all comparisons the ranking provided by the pseudo metric is superior to the FID metric and mimics the monotonically decreasing behavior of the true metric.

\section*{CRediT authorship contribution statement}
\textbf{Firas Al-Hindawi:} Conceptualization, Methodology, Software, Writing - Original Draft, Writing - Review \& Editing. \textbf{Md Mahfuzur Rahman Siddiquee:} Conceptualization, Methodology, Software, Writing - Review \& Editing. \textbf{Teresa Wu:} Conceptualization, Methodology, Writing - Original Draft, Writing - Review \& Editing, Resources, Supervision, Project administration. \textbf{Han Hu:} Conceptualization, Writing - Original Draft, Writing - Review \& Editing, Data Curation. \textbf{Ying Sun} Project administration, Supervision.

\section*{Declaration of Competing Interest}
The authors declare that they have no known competing financial interests or personal relationships that could have appeared to influence the work reported in this paper

% \section*{Acknowledgement}
% Support for this work was provided in part by the US National Science Foundation under Grant No. CBET-1705745.

% \bibliographystyle{ieeetr}
\bibliography{references.bib}

\end{document}